%% file: top.tex
\pdfoutput=1
\documentclass[10pt,twocolumn,letterpaper]{article}

\usepackage{iccv}
\usepackage{times}
\usepackage{epsfig}
\usepackage{graphicx}
\usepackage{amsmath}
\usepackage{amssymb}
\usepackage[ruled,vlined,linesnumbered]{algorithm2e}

\usepackage{multirow}
\usepackage{color}
\usepackage{booktabs}
\usepackage{multirow}
\usepackage{subcaption}
\usepackage{gensymb}
\usepackage{bm}

\graphicspath{{./figs/}}

\usepackage{soul}
\usepackage{tabularx}
\usepackage[title]{appendix}
\usepackage{titling}

\newcommand{\norm}[1]{\left\| {#1} \right\|}

\newcommand{\R}{\ensuremath \mathbb{R}}

\newcommand{\G}{\mathcal{G}}

\renewcommand{\P}{\mathcal{P}}

\newcommand{\CC}{\mathcal{C}}

\newcommand{\D}{\mathcal{D}}

\renewcommand{\epsilon}{\varepsilon}

\def\utilde#1{\mathord{\vtop{\ialign{##\crcr
$\hfil\displaystyle{#1}\hfil$\crcr\noalign{\kern1.5pt\nointerlineskip}
$\hfil\tilde{}\hfil$\crcr\noalign{\kern1.5pt}}}}}

\newcommand{\hide}[1]{}

\usepackage[breaklinks=true,bookmarks=false]{hyperref}

\iccvfinalcopy 

\begin{document}

\title{DAGMapper: Learning to Map by Discovering Lane Topology}

\author{
	Namdar Homayounfar$^{1,2}$ \quad Wei-Chiu Ma$^{1,3}$\\
	Justin Liang$^{1}$ \quad Xinyu Wu $^{1}$ \quad Jack Fan $^{1}$ \quad Raquel Urtasun$^{1,2}$\\
	$^{1}$Uber Advanced Technologies Group \quad $^{2}$University of Toronto \quad $^{3}$ MIT\\
	\small\texttt{namdar.homayounfar@mail.utoronto.ca, weichium@mit.edu} \\
	\small\texttt{justin.j.w.liang@gmail.com, xy.wu91@gmail.com, me@jackfan.com} \\
	\small\texttt{urtasun@cs.toronto.edu}
}
\date{}

\maketitle

\input{abs}

\input{intro}

\input{related}

\input{model}

\input{experiments}

\input{conclusion}

{\small
\bibliographystyle{ieee_fullname}
\bibliography{top}
}

\onecolumn
\title{Appendix-- DAGMapper: Learning to Map by Discovering Lane Topology}
\author{\vspace{-5ex}}
\date{\vspace{-5ex}}
\maketitle

\begin{appendices}
	\input{supp_model}

	\input{supp_training}
	\input{qual}

\end{appendices}

\end{document}

%% file: abs.tex
 
\begin{abstract}

One of the fundamental challenges to scale self-driving  is being able to create accurate high definition maps (HD maps) with low cost. 
Current attempts to automate this process   typically focus on simple scenarios, estimate independent maps per frame or do not have the level of precision required by modern self driving vehicles. 
In contrast, in this paper we focus on  drawing the lane boundaries of complex highways with many lanes that contain topology changes due to forks and merges. 
Towards this goal, we formulate the problem as inference in a directed acyclic graphical model (DAG), where the nodes of the graph encode geometric and topological properties of the local regions of the lane boundaries. Since we do not know a priori the topology of the lanes, we also infer the DAG topology (i.e., nodes and edges) for each region. 
We demonstrate the effectiveness of our approach on  two major North American Highways in two different states and show high precision and recall as well as  89\% correct topology.  
\end{abstract}

%% file: intro.tex
\section{Introduction}

Self-driving vehicles are equipped with a plethora of sensors, including GPS, cameras, LiDAR, radar and ultrasonics. This allows the vehicle to see with a field of view of 360 degrees, potentially having super human capabilities. 
Despite decades of research, building reliable solutions that can handle the complexity of the real world is still an open problem. 

Most modern  self-driving vehicles utilize a high definition map (HD map), which contains information about the location of lanes, their types, crosswalks, traffic lights, rules at intersections, etc. 
This information is very accurate, typically with error on the order of a few centimeters. It can thus help localization \cite{ma2017find,barsan2018learning,ma2019exploit}, perception \cite{yang2018hdnet,liang2019multi,casas2018intentnet}, motion planning \cite{zeng2019end}, and even simulation \cite{fang2018simulating}.
Building these maps is however one of the main obstacles to building self driving cars at scale. The typical process is very cumbersome, involving a fleet of vehicles passing multiple times across the same area to build accurate geometric representations of the world.
Annotators then  label the location of all these semantic landmarks by hand from a bird's eye view representation of the world. 
While building the geometric map is time consuming, manual annotation is costly in terms of capital. 

\begin{figure}[t]

	\centering
	\includegraphics[width=\linewidth]{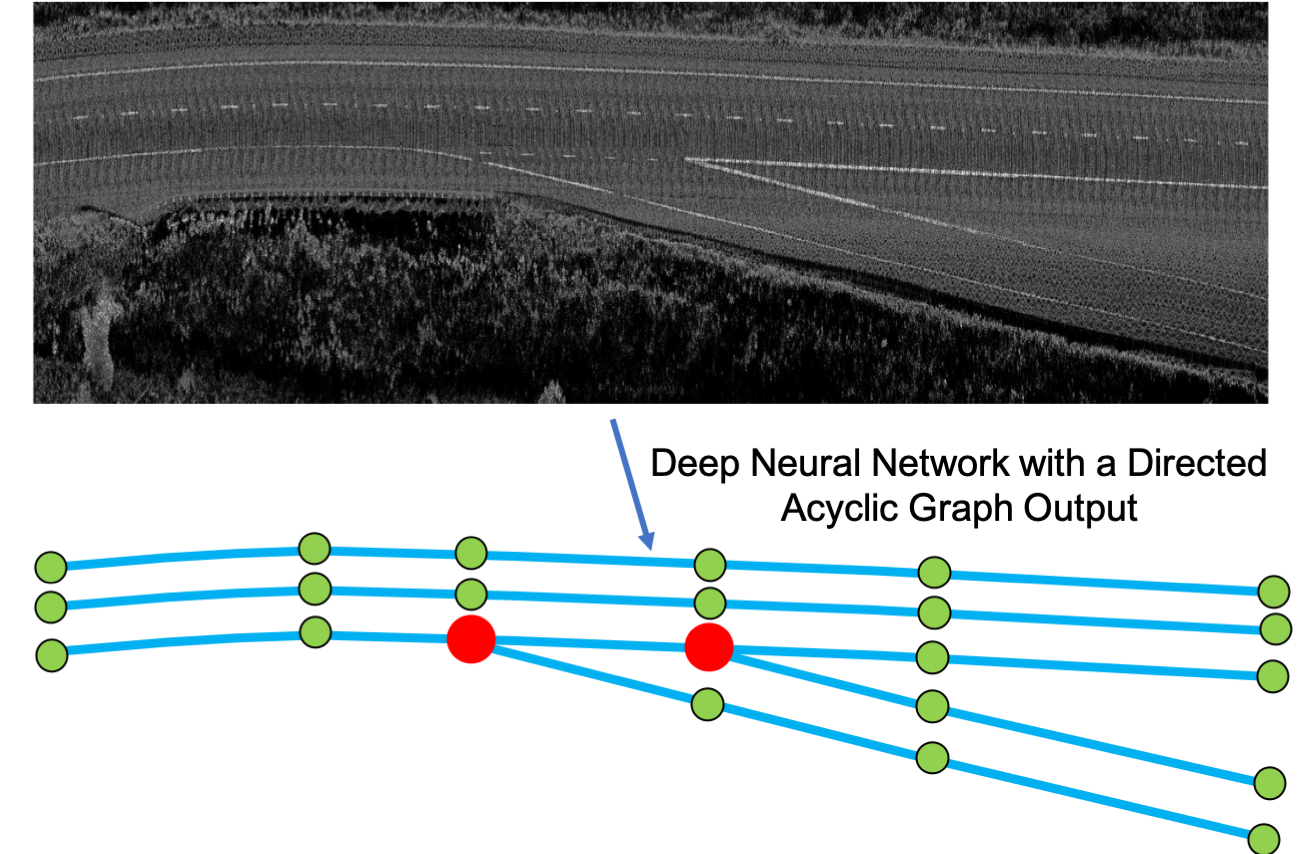}
	\caption{The input to our model is an aggregated LiDAR intensity image and the output is a DAG of the lane boundaries parametrized by a deep neural network. }
	\label{fig:qual-results}
\end{figure}

In this paper, we tackle the problem of automatically creating HD maps of highways that are consistent over large areas. Unlike the common practice in the industry, we aim to map the whole region from a \emph{single} pass of the vehicle.
Towards this goal, we first capitalize on the LiDAR mounted on our self-driving vehicles to build a bird's eye view (BEV) representation of the world. We then exploit a deep network to extract the exact geometry and topology of the underlying lane network.

\begin{figure*}[t!]
	\vspace{-0.5cm}
	\centering
	\includegraphics[width=0.8\linewidth,trim={0mm 75mm 30mm 0mm},clip]{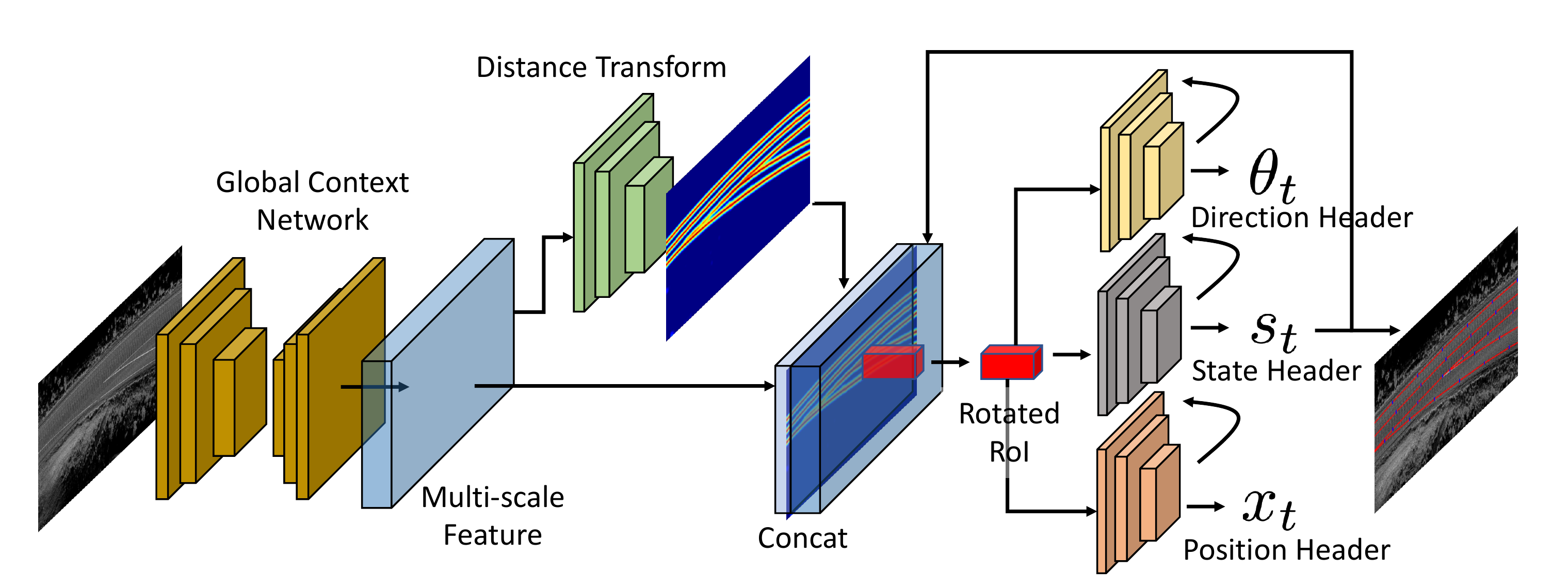}
	\caption{Network Architecture: A deep global feature network is applied on the LiDAR intensity image followed by three recurrent convolutional headers that parametrize a DAG of the lane boundaries. DT image is used for initialization of the lane boundaries as well as recovery from mistakes.}
	\label{fig:network}
\end{figure*}

The main challenge of this task arises from the fact that  highways contain complex topological changes due to forks and merges, where a lane boundary splits into two or two lane boundaries merge into one. We address these challenges by formulating the problem as inference in a directed acyclic graphical model (DAG), where the nodes of the graph encode geometric and topological properties of the local regions of the lane boundaries. Since we do not know a priori the lane topology, we also infer the DAG topology (i.e., nodes and edges) for each region. We leverage the power of deep neural networks to represent the conditional probabilities in the DAG. We then devise a simple yet effective greedy algorithm to incrementally estimate the graph as well as the states of each node. We learn all the weights of our deep neural network in an end-to-end manner. We call our method DAGMapper.

We demonstrate the effectiveness of our approach on two major North American Highways in two different states 1000 km apart. The dataset consists of LiDAR sensory data and the corresponding lane network topology  on areas of high complexity such as forks and merges. 
Our approach achieves a precision and recall of 89\% and 88.7\% respectively at a maximum  of 15 cm away from the actual location of the lane boundaries.  Moreover, our approach obtains the correct topology 89\% of the time. Finally, we showcase  the strong generalization of our approach by training on a highway and testing  on another belonging to a different state.  Our experiments show that no domain adaptation is required.

%% file: related.tex
\section{Related Work}

\vspace{-0.2cm}

\paragraph{Road and lane detection:}
Detecting the lane markings \cite{vivacqua2017self,behrendt2017deep}, inferring the lane boundaries \cite{bai2018,wang2018lanenet,hou2019learning} as well as finding the drivable surface \cite{franke1996fast,badino2007free,teichmann2018multinet} on the road is of utmost importance for self-driving vehicles.
Many methods  have been proposed to tackle this problem. The authors in  \cite{tan2006color, lieb2005adaptive, paz2015variational, alvarez2011road,kong2010general,cheng2006lane,wedel2009b, alvarez2012road, kuhnl2012spatial} use appearance and geometric cues to detect the road in an unsupervised or semi-supervised fashion, while \cite{mohan2014deep,levi2015stixelnet, yao2015estimating} apply graphical models to estimate  free space. With the advent of modern deep learning \cite{lecun2015deep}, a new line of research for lane marking detection has been established. In \cite{lee2017vpgnet}, the authors employ vanishing points and train a neural network that detects lane boundaries. \cite{ghafoorian2018gan} uses generative adversarial networks \cite{goodfellow2014generative} to further refine the detection and obtain thinner segmentations. The authors in \cite{azimi2018aerial} design a symmetric CNN enhanced by wavelet transform in order to segment lane markings from aerial imagery. 
In \cite{bai2018}, camera and LiDAR imagery are employed to predict a dense representation of the lane boundaries in the form of a thresholded distance transform. In \cite{Neven_2018}, the authors treat lane detection as an instance segmentation task where deep features corresponding to lane markings are clustered.

\vspace{-0.2cm}

\paragraph{Road network extraction:}
Leveraging aerial and satellite imagery for the task of road network extraction goes back many decades \cite{Fortier2002SurveyOW, richards_2013}. In the early days, researchers mainly extracted the road network topology by iteratively growing a graph using simple heuristics on spectral features of the roads \cite{simonett1970use,Bajcsy1976ComputerRO}. Recently, deep learning and more advanced graph search algorithms have been leveraged  \cite{Mena2005AnAM, Mnih2010LearningTD,Mnih2012LearningTL,Marmanis2016ClassificationWA,marmanis2016semantic, mattyus2017deeproadmapper, ventura2018iterative,bastani2018roadtracer} to extract the road network from aerial imagery more effectively. For instance, \cite{li2018polymapper} extract building and road network graphs directly in the form of polygons from aerial imagery.  \cite{mattyus2015enhancing, mattyus2016hd} enhance Open Street Maps with lane markings, sidewalks and parking spots by applying graphical models on top of deep features. \cite{sun2019leveraging} leverages GPS data to enhance road extraction from aerial imagery. \cite{Batra_2019_CVPR} predicts both the orientation and the semantic mask of the road to obtain topologically correct and a connected road network from aerial imagery.  One however should note that these methods perform road network extraction and semantic labeling at a \emph{coarse} scale. While they can be beneficial for routing purposes, they lack the required resolution for self driving applications. 

\vspace{-0.2cm}

\paragraph{High Definition maps:}
Creating HD maps that have centimeter level accuracy is crucial for the safe navigation of  self-driving vehicles.
Recently, researchers have been devoted to generating HD maps automatically from various sensory data \cite{kammel2008lidar}. For example, the authors in \cite{liang2018end} extract crosswalk polygons from top-down LiDAR and camera imagery. \cite{liang2019road,homayounfar2018hierarchical} employ deep convolutional networks to extract road and lane boundaries in the form of polylines.
These lines of mapping work are similar to instance segmentation methods \cite{PolyRNN,acuna2018efficient,ling2019fast} where structured representations of objects such as polygons are obtained. The algorithms are thus amenable to an annotator in the loop.

At a high level, our work shares similarities  with \cite{homayounfar2018hierarchical},  which  predicts lane boundaries from a top down BEV LiDAR image by exploiting  a recurrent convolutional neural network. However, \cite{homayounfar2018hierarchical} cannot handle changes of topology of lane boundaries and has baked in notions of lane boundary ordering which we address in this paper.

%% file: model.tex

\section{Learning to Map }

Our goal is to create HD maps directly from 3D sensory data. More specifically, the input to our system is a BEV aggregated LiDAR intensity image $\mathcal{D}$ and the output is a collection of structured polylines corresponding to the lane boundaries. Rather than focusing on local areas, we attempt to extract the exact topology and geometry of the  lane network on a long stretch of the highway. 
 This is extremely challenging as highways contain complex topological changes (\eg, forks and merges) and the area of coverage is very large.
Towards this goal, we formulate the underlying geometric topology as a directed acyclic graph and present a novel recurrent convolutional network that can effectively extract such topology from LiDAR data. We start our discussion by describing our output space formulation in terms of the DAG. Next, we explain how we exploit neural networks to parameterize each conditional probability distribution. Finally, we describe how inference and learning are performed.

\subsection{Problem Formulation and DAG Extraction}

We draw our inspirations from how humans create lane graphs when building maps. Given an initial vertex, the annotators first trace the lane boundary with a sequence of clicks that respect the  local geometry. Then, if there is a change of topology during the process, the annotators identify such points and annotate correspondingly. For instance, if the click reaches the end of the road, one simply stops; if there is a fork, one simply needs to create another branch from the click. By using this simple approach, the annotators can effectively label the road network in the form of a DAG. In this work, we design an approach to road network DAG discovery that mimics such an annotation process.

\begin{figure*}[t!]
	\vspace{-0.5cm}
	\centering
	\includegraphics[width=0.9\linewidth]{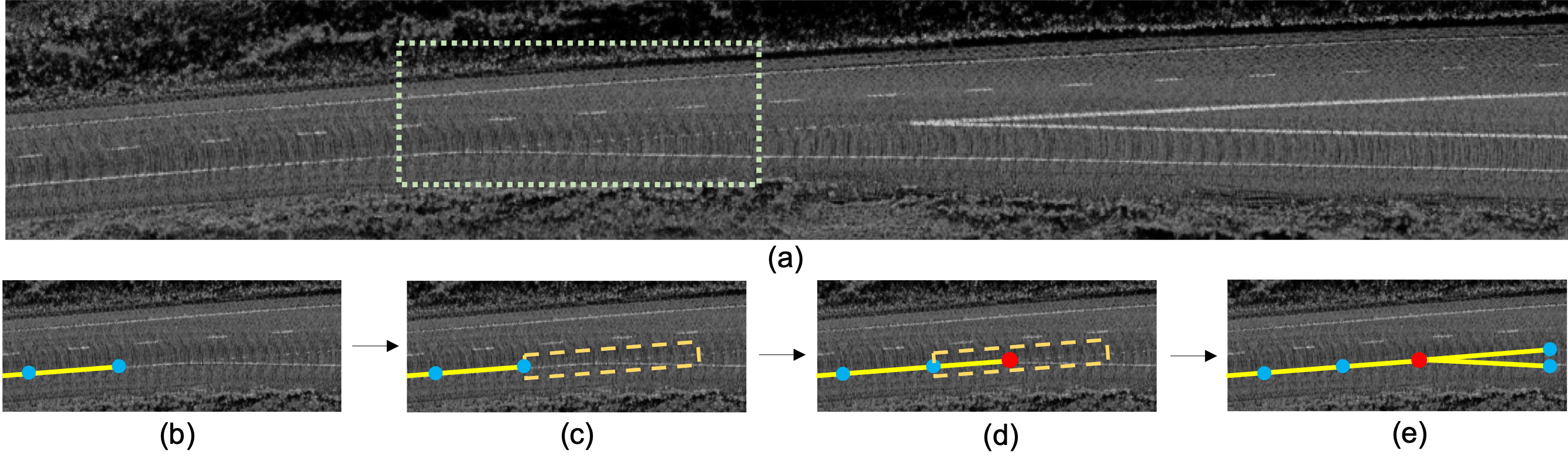}
	\caption{(a) The BEV aggregated LiDAR image $\mathcal{D}$ of an area of the highway with a fork. The green box shows a zoomed in area for visualization of the inference process. (b) The current subgraph of the inferred bottom lane boundary. (c) The direction $\theta_i $of the new ROI (yellow box) along the lane boundary is predicted. (d) A new node is predicted with position $x_i$ (red dot) within the ROI with the \emph{fork} state $s_i$. (e) Two lane boundaries emanate from the fork node and the process continues. }
	\label{fig:lidar_img}
\end{figure*}

\begin{algorithm}[tb]
	\SetAlgoLined
	\SetKwInOut{Input}{Input}\SetKwInOut{Output}{Output}
	\Input{Aggregated point clouds, initial vertices $\{v_{\text{init}} = (\theta_{\text{init}}, x_{\text{init}}, s_{\text{init}})\}$}
	\Output{Highway DAG Topology}
	Initialize $queue~Q$ with vertices $\{v_{\text{init}}\}$\;
	\While{Q not empty}{
		$v_i \leftarrow Q.pop()$\;
		$i \leftarrow \mathcal{C}(i)$\;
		\While{$s_{\mathcal{P}(i)}$ not Terminate}{
			$\theta_i \leftarrow \arg\max~p(\theta_i | \theta_{\P(i)}, s_{\P(i)}, x_{\P(i)})$\;
			$x_i \leftarrow \arg\max~p(x_i| \theta_i, s_{\P(i)}, x_{\P(i)})$\;
			$s_i \leftarrow \arg\max~p(s_i| \theta_i, s_{\P(i)}, x_{\P(i)})$\;
			\If{$s_i = $ Fork}{
				$Q.insert(v_{i})$\;
			}
			$i \leftarrow \mathcal{C}(i)$\;
		}
	}
	\caption{Deep DAG Topology Discovery}
	\label{alg:dag-discovery}
\end{algorithm}

More formally, let $G = (V, E)$ be a DAG, where $V$ and $E$ denote the corresponding set of nodes and edges  defining the  topology. Each node $v_i=  (x_i,\theta_i, s_i)$ in the DAG encodes geometric and topological properties of a local region of the lane boundary. We further use $v_{\P(i)}$ and $v_{\CC(i)}$ respectively to denote the parent and the child of the node $v_i$.
The geometric component $x_i$ denotes the position of the vertex in global coordinates and $\theta_i$ refers to the turning angle at the previous vertex position $x_{\P(i)}$. 
The state component $s_i$ is a categorical random variable that denotes whether to continue the lane boundary without any change of topology, to spawn an extra vertex for a new lane boundary at a fork, or to terminate the lane boundary (at a merge). 
Therefore, if $s_i$ is a fork, then $|\mathcal{C}(i)| = 2$; if $s_i$ is stop at the lane end of the lane boundary, then $|\mathcal{C}(i)| = 0$.

Given the aggregated BEV LiDAR data  $\mathcal{D}$, our goal is to find the Maximum A Posteriori (MAP) over the space of all possible graphs $\G$:

\begin{align}
G^{*} = \arg\max_{G \in \mathcal{G}} p(G|\mathcal{D}).
\end{align}

As $G$ is a DAG in our case, the joint probability distribution $p(G|\mathcal{D})$ can be factorized into:

\begin{align}
p(G| \D) = \prod_{v_i \in V} p(v_{i} | v_{\P(i)}, \D ),
\label{eq:factorization}
\end{align}
where each conditional probability further decomposes into geometric and topological components:
\begin{align}
p(v_{i} | v_{\P(i)}, \D ) = &p(\theta_i | \theta_{\P(i)}, s_{\P(i)}, x_{\P(i)}, \D) \notag\\
	&\times p(x_i| \theta_i, s_{\P(i)}, x_{\P(i)}, \D) \notag\\
	&\times p(s_i | \theta_i, s_{\P(i)}, x_{\P(i)}, \D).
\label{eq:component}
\end{align}
This conditional distributions are modeled with deep neural networks in order to handle the complexities of the real world. Fig. \ref{fig:lidar_img} shows an example BEV image $\D$ and the inference process of a lane boundary where there is a fork. 

We provide in the next section an in depth discussion of the networks.

\vspace{-0.4cm}

\paragraph{Topology Extraction:} Unfortunately, the space of $\G$ is exponentially large. It is  thus computationally intractable to consider all possible DAGs.

To tackle this issue, we design an approach that \emph{greedily} constructs the graph and computes the geometric and topological properties of each vertex. 
Specifically, we take the $\arg\max$ of each component in Eq. (\ref{eq:component}) to obtain the most likely topological and geometric states for each node $v_i$.
Given a parent node $v_{\P(i)}$, we first predict along which direction $\theta_i$ the node $v_i$ would be. Then we attend to a local region of interest along the lane boundary specified by $\theta_{i}$ and predict the vertex position $x_{i}$ and the local state $s_{i}$.
The state specifies that either (1) there is no change of topology and we continue as normal, (2) there is a fork and we continue the current lane boundary as normal but at the same time spawn a new vertex for a new lane boundary, or (3) the current lane boundary has terminated at the merge, and hence the node $v_i$ has no child. By iterating through the procedure, we are able to obtain the structure of the DAG as well as their geometric positions. Our structure discovery algorithm is summarized in Alg. \ref{alg:dag-discovery}.  We next describe the backbone network that we used to extract the context features and the header networks that approximate the geometrical and topological probability distribution.

\subsection{Network Design}
In this work, we exploit neural networks to approximate the DAG probability distributions described in the previous section, also  shown in  Alg. \ref{alg:dag-discovery}.
At a high level, as shown in Fig. \ref{fig:network}, we first exploit a \emph{global feature network} to extract multi-scale features and a \emph{distance transform network} to encode explicit lane boundary information. The features are then passed to the \emph{direction header} to predict  a rotated Region of Interest (RoI) along the lane boundary. Finally, the \emph{position header} and the \emph{state header} condition on the features within the RoI  predict the position and state of the next vertex. We now present the high level details of each head and refer the reader to the appendix for a description of the  full architecture. 

\paragraph{Global Feature Network:} The aim of this network is to build features for the header networks to draw upon. Since the changes of topology are usually very gradual, it will be very difficult to infer them if the features capture merely local observations or the network has a small receptive field (see Fig. \ref{fig:lidar_img}). For instance, at a fork, a lane boundary gradually splits into two so that vehicles have enough time to exit the highway and slow down in a safe manner. Similarly, at a merge, two lanes become one gradually over a long stretch so that vehicles accrue speed and enter the highway smoothly without colliding with the ongoing flow of traffic. 

The input to this network is a BEV projected aggregated LiDAR intensity image $\mathcal{D} \in \R^{1\times H\times W}$ at 5 cm per pixel. Our images are of dimension 8000 pixels in width by maximum 1200 in height corresponding to 400m by maximum 60m in the direction of  travel of the mapping vehicle as shown in Fig. \ref{fig:lidar_img}. As such, to better infer the state of each vertex, one must exploit larger contextual information. Following \cite{homayounfar2018hierarchical}, we adopt an encoder-decoder architecture \cite{linknet} built upon a feature pyramid network \cite{lin2016feature} that encodes the context of the lane boundaries and the scene. The bottom-up, top-down structure enables the network to process and aggregate multi-scale features; the skip links help preserve spatial information at each resolution. The output of this network is a feature map $F \in \R^{C\times \frac{H}{4}\times \frac{W}{4}}$.

\begin{table*}[t!]
	\vspace{-0.5cm}
	\centering
\begin{tabular}{c|c|c|c|c||c|c|c|c||c|c|c|c||}
\cline{2-13}
 & \multicolumn{4}{c||}{Precision at (px)} & \multicolumn{4}{c||}{Recall at (px)} & \multicolumn{4}{c||}{F1 at (px)} \\ \cline{2-13} 
 & 2 & 3 & 5 & 10 & 2 & 3 & 5 & 10 & 2 & 3 & 5 & 10 \\ \hline
\multicolumn{1}{|c||}{DT baseline \cite{bai2018}} 		& 68.1 & 87.9 & \bf 96.3 & \bf 98.0
								& 65.8 & 85.0 & 93.5 & \bf 96.2
								& 66.9 & 86.4 & \bf 94.9 & \bf 97.0 \\ \hline

\multicolumn{1}{|c||}{HRAN \cite{homayounfar2018hierarchical} with GT init} 		
&55.6 & 71.0 & 84.0 & 90.9
&45.9 & 58.1 & 68.3 & 74.0 
&50.1 & 63.7 & 75.1 & 81.4 \\ \hline

\multicolumn{1}{|c||}{Ours} 	
&\bf 76.4 & \bf 89.0 & 94.6 & 96.6 
&\bf 76.2 & \bf 88.7 & \bf 94.2  & 96.1
&\bf 76.3 & \bf 88.8 & 94.4 & 96.3  \\ \hline

\end{tabular}
	\caption{Comparison to the baselines from \cite{bai2018,homayounfar2018hierarchical}. We highlight the precision, recall and F1 metrics at thresholds of 2, 3, 5 and 10 px (5cm/px).}
	\label{tab:baseline}
\end{table*}

\vspace{-0.1cm}

\paragraph{Distance Transform Network:} The distance transform (DT) and more specifically the thresholded inverse DT has been proven to be an effective feature for mapping \cite{liang2018end,liang2019road}. It encodes at each point in the image the relative distance to the closest lane boundary. As such, we employ a header that consists of a sequence of residual layers that takes as input the global feature map $F$ and outputs a thresholded inverse DT image $D \in R^{1\times \frac{H}{4}\times \frac{W}{4}}$.     
We use this DT image $D$ for three purposes: 1) We stack it to the global feature map $F$ as an additional channel and feed it to the other headers. Our aim is to use $D$ as a form of attention on the position of the lane boundaries. 2) We threshold, binarize and skeletonize $D$ and obtain the endpoints of the skeleton as the initial vertices of the graph. 3) After inferring the graph using Alg. \ref{alg:dag-discovery}, we draw the missed lane boundaries by initializing points on the regions of the $D$ that are not covered by our graph. 

\vspace{-0.1cm}

\paragraph{Direction Header:} This network serves as an approximation to $p(\theta_i | \theta_{\P(i)}, s_{\P(i)}, x_{\P(i)})$. Given the geometrical and topological information of the parent vertex, this header predicts the direction of the rotated RoI along the lane boundary where the current vertex lies. 
The input to this header is a bilinearly interpolated axis aligned crop from the concatenation of $F$, $D$ centered at $x_{\P(i)}$ and the channel-wise one hot encoding of the state $s_{\P(i)}$. 
At a fork or a merge, the two lane boundaries are very close to each other. Conditioning on the state vector signals to the header to predict the correct direction corresponding to its lane boundary. This input is fed into a simple convolutional RNN that outputs a direction vector of the next rotated ROI. We employ an RNN to encode history of the previous directions and states.

\vspace{-0.1cm}

\paragraph{Position Header:} This network can be viewed as an approximation to $p(x_i | \theta_{i}, s_{\P(i)}, x_{\P(i)})$. Given the state and position of the previous vertex, the network predicts a probability distribution over all possible positions within the rotated ROI along the lane boundary generated by the direction header. This RoI is bilinearly interpolated from the concatenation of $F$, $D$ and channel-wise one-hot encoding of $s_{\P(i)}$. After the interpolation, we upsample the region to the original image dimension and pass it to a convolutional recurrent neural network (RNN). The output of the RNN is fed to a lightweight encoder-decoder network that outputs a softmax probability map of the position $x_i$ that is mapped to the global coordinate frame of the image.

\paragraph{State Header} This header can be regarded as an approximation to $p(s_i | \theta_{i}, s_{\P(i)}, x_{\P(i)})$, which infers the local topological state of the lane boundaries. Specifically, the network predicts a categorical distribution specifying whether to continue drawing normally, to stop drawing, or to fork a new lane boundary as we have arrived to  a merge. The input to this model is the same rotated RoI of the position header. A convolutional RNN is employed to encode the history and the output is the softmax probability over the three states.

\subsection{Learning}
We employ a multi task objective to provide supervision to all the different parts of the model. Since all the components are differentiable we can   learn our model end-to-end. In particular, similar to \cite{homayounfar2018hierarchical} we employ the symmetric Chamfer distance to match each GT polyline $Q$ to its prediction $P$ : 
\begin{align}
L(P, Q) &= \sum_{i}\min_{q \in Q}{ \norm{p_i - q}_2} + \sum_{j} \min_{p \in P}{ \norm{p - q_j}_2} \label{eq:poly_loss}
\end{align}
where $p$ and $q$ are the densely sampled coordinates on polylines $P$ and $Q$ respectively. To learn the topology states, we use multi label focal loss with a slight modification; Rather than taking the mean of all the individual losses, we add them up and divide by the sum of the focal weights. 
Here, the intuition is that the wrong predictions are more emphasized and are not dampened by the over-sampled class corresponding to the \emph{normal} state.  Finally, we employ cosine similarity loss and the $\ell_2$ loss to learn the directions and the distance transform. We refer the reader to the appendix for the full training and implementation details.

%% file: experiments.tex

\section{Experimental Evaluation}

\begin{table*}[t!]
	\vspace{-0.5cm}
	\centering
\begin{tabular}{cccc||c|c|c|c||c|c|c|c||c|c|c|c||}
\cline{5-16}
 &  &  & & \multicolumn{4}{c||}{Precision at (px)} & \multicolumn{4}{c||}{Recall at (px)} & \multicolumn{4}{c||}{F1 at (px)} \\ \hline
\multicolumn{1}{|c|}{G} & \multicolumn{1}{c|}{M} & \multicolumn{1}{c|}{R} & S & 2 & 3 & 5 & 10 & 2 & 3 & 5 & 10 & 2 & 3 & 5 & 10 \\ \hline
\multicolumn{1}{|c|} {-} & \multicolumn{1}{c|}{-} &  \multicolumn{1}{c|}{-} & {\checkmark} 

&75.4  &87.4 & 91.6 & 93.5
&62.1  &72.0 & 75.6 & 77.4
&68.1 & 79.0 & 82.8 & 84.7
\\ \hline

\multicolumn{1}{|c|}{\checkmark} & \multicolumn{1}{c|}{-} &  \multicolumn{1}{c|}{-} &  {\checkmark} 

&74.1 & 86.6 & 93.0 & 95.9
&65.6 & 76.5 & 82.0 &84.5
&69.5 & 81.2 & 87.2  & 89.8

\\ \hline

\multicolumn{1}{|c|}{\checkmark} & \multicolumn{1}{c|}{-} &  \multicolumn{1}{c|}{\checkmark} & {\checkmark} 

&73.3 & 85.9 & 92.6 & 95.4
&72.8 &85.2 & 91.6  &94.4
&73.1 & 85.6 & 92.1 & 94.9

\\ \hline

\multicolumn{1}{|c|}{\checkmark} & \multicolumn{1}{c|}{\checkmark} &  \multicolumn{1}{c|}{\checkmark} &  - 

& 69.8 &83.0 & 90.6 & 93.4
& 70.2 & 83.5 &91.2 & 94.2
& 70.0 & 83.3 & 90.9 & 93.8

\\ \hline

\multicolumn{1}{|c|}{\checkmark} & \multicolumn{1}{c|}{\checkmark} &  \multicolumn{1}{c|}{-} &  {\checkmark} 

& \bf 77.0 & \bf 89.5 & \bf 95.1 & \bf 96.9
& 75.0 & 87.2 & 92.4 & 94.2
& 76.0 & 88.3 & 93.7 & 95.5

\\ \hline

\multicolumn{1}{|c|}{\checkmark} & \multicolumn{1}{c|}{\checkmark} &  \multicolumn{1}{c|}{\checkmark} &  {\checkmark} 

& 76.4 &  89.0 & 94.6 & 96.6 
&\bf 76.2 & \bf 88.7 & \bf 94.2  & \bf 96.1
&\bf 76.3 & \bf 88.8 & \bf 94.4 & \bf 96.3  \\ \hline


\end{tabular}
	\caption{We evaluate the contribution of the different components of our model. In particular, we assess the global feature network (G), the conv-RNN (M), the notion of state (S), and recovery using the DT image (R) in case of missing a lane boundary. }
	\label{tab:ablation}

\end{table*}

\begin{table}[]
	\vspace{-0.3cm}
	\centering
	\begin{tabular}{c|c|c|c|c|}
		\cline{2-5}
		& 2 px & 3 px & 5 px & 10 px \\ \hline
		\multicolumn{1}{|c|}{Precision} & 57.4 & 72.1 & 83.4 & 88.6        \\ \hline
		\multicolumn{1}{|c|}{Recall}    &  53.0 & 66.2 & 76.1 & 80.9       \\ \hline
		\multicolumn{1}{|c|}{F1}        &   55.1 & 69.0 & 79.5 & 84.6    \\ \hline
	\end{tabular}
	\caption{Generalization to a new highway: we report precision, recall and F1 metrics of training on all the images of one highway and testing on a completely new highway.}
	\label{tab:sf}
\end{table}

\paragraph{Dataset:}
Our dataset consists of LiDAR point clouds captured by driving a mapping vehicle multiple times through a major North American Highway. For each run of the vehicle, we aggregate the point clouds using IMU odometry data in a common coordinate frame with an arbitrary origin. In our setting of offline mapping, the aggregation of point clouds in a local area is taken from all the past as well future LiDAR sweeps that hit that area. Next, we project the point clouds onto BEV and rasterize at 5 cm per pixel resolution by taking the intensity of the returned point with the lowest elevation value. Since we are interested in HD mapping of lane boundaries that fall on the surface of the road, by taking the lowest elevation point we aim to filter out the LiDAR return from other moving vehicles. 
  
We are interested in mapping difficult areas on the highway such as forks and merges. As such we locate where these interesting topologies occur and crop a rectangular region of 8000 pixels by a maximum of 1200 pixels with the long side of the rectangle perpendicular to the trajectory of the mapping vehicle. Note that the height of the image varies in each example depending on the curvature of the road. This crop corresponds to a 400m by maximum 60m region.   

We annotate the ground truth by drawing polyline instances associated to each lane boundary. At a fork we assign a new polyline to the new lane boundary that appears. In total we create 3000 images from 68 unique fork/merge areas on the highway. 

Since our aim is scalable high definition mapping of highways, we evaluate the effectiveness of our method on a different highway in a different geographical region. As such we drive our mapping vehicle multiple times in this new highway and collect and annotate 336 images from 114 unique fork/merge areas on the highway. We use this smaller dataset only for evaluation. 

To create our train, val and test splits, we first divide our 68 fork/merge areas randomly with the ratio of 70:15:15 and place the images of each area in the corresponding split. This way the same fork/merge area is not present in different splits. To gain statistical significance in our reported metrics, we repeat this procedure three time to obtain three cross validation splits. As such, for baseline comparisons and ablation studies, we train and test each model three times. 

\begin{figure*}[t]
	\vspace{-0.5cm}
	\centering
	\setlength{\tabcolsep}{1pt}
	\begin{tabular}{cc}

		\raisebox{23px}{\rotatebox{90}{GT \hspace*{4em} Prediction}}
		\includegraphics[width=0.90\linewidth, trim={0 0 128px 0},clip]{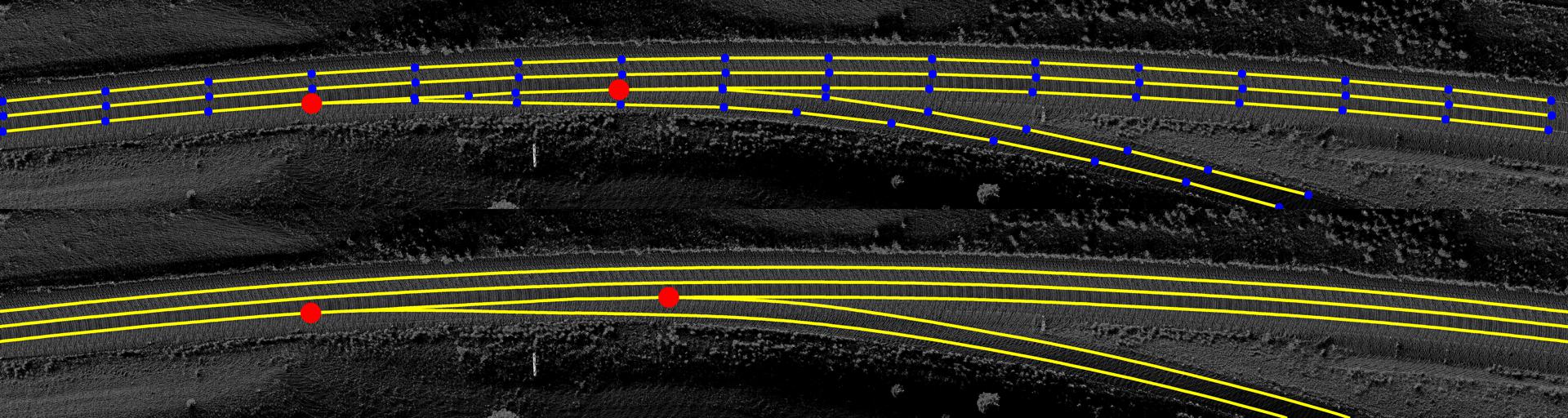} \\ 
		\raisebox{23px}{\rotatebox{90}{GT \hspace*{4em} Prediction}}
		\includegraphics[width=0.90\linewidth, trim={0 0 128px 0},clip]{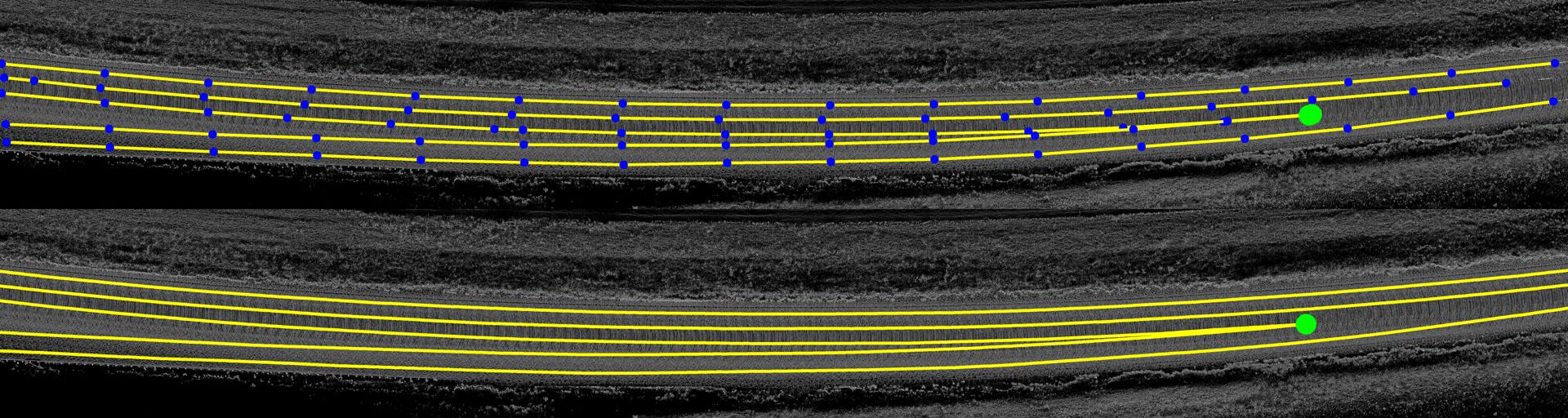}  \\

		\raisebox{23px}{\rotatebox{90}{GT \hspace*{4em} Prediction}}
		\includegraphics[width=0.90\linewidth, trim={0 0 128px 0},clip]{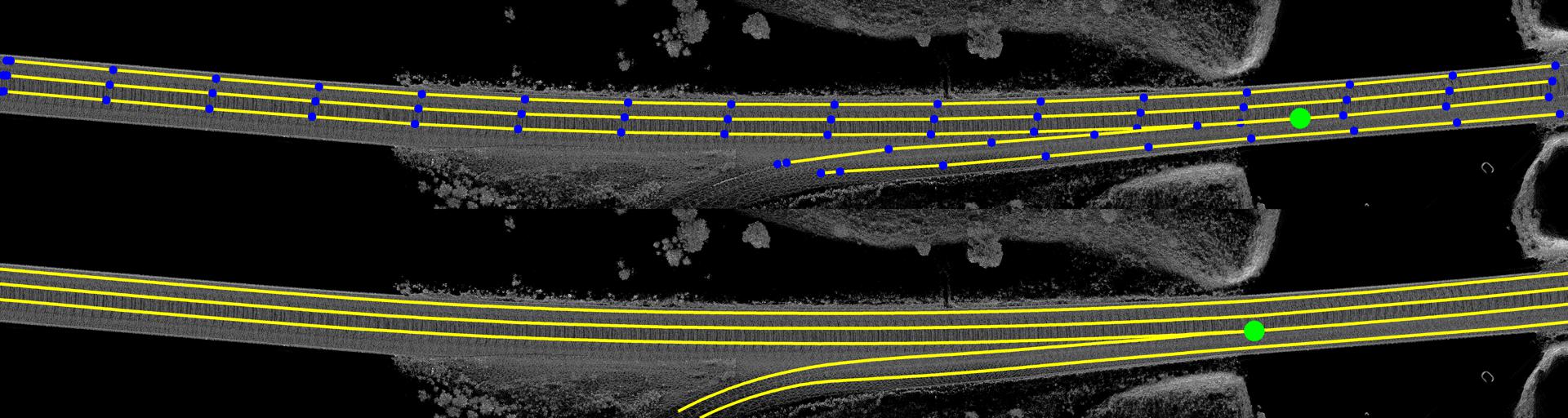} \\									        
		\raisebox{23px}{\rotatebox{90}{GT \hspace*{4em} Prediction}} 
		\includegraphics[width=0.90\linewidth, trim={0 0 128px 0},clip]{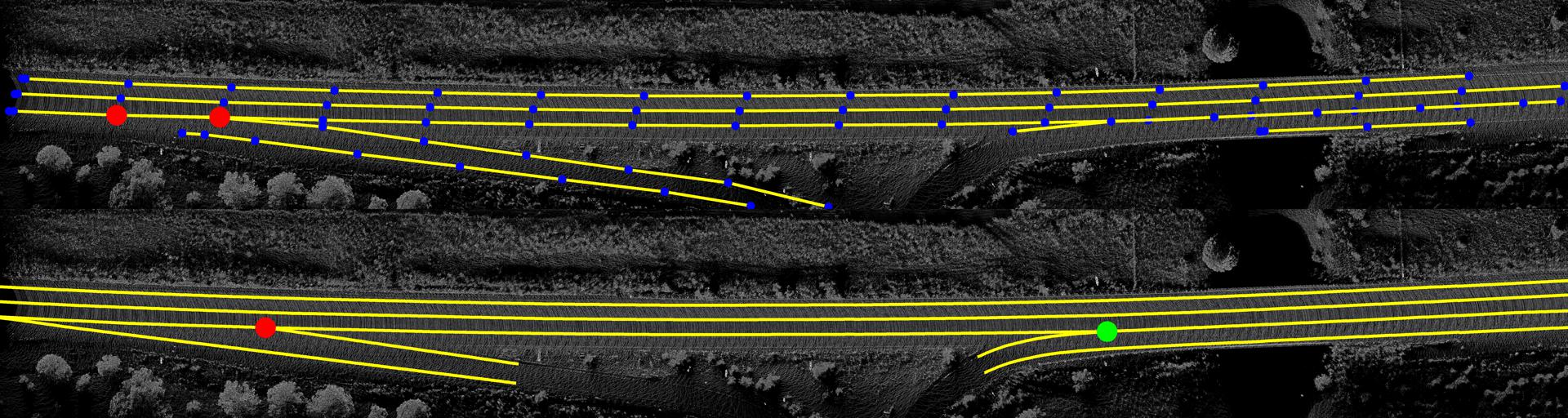}  \\

	\end{tabular}

	\caption{Qualitative results: Rows 1-3 showcase the effectiveness of our method. Row 4 is a failure case where two lane boundaries are bottom are drawn partially.}
	\label{fig:qual}
\end{figure*}

\paragraph{Baselines:}

Since there are no baselines in the literature that perform offline mapping on long stretches of the highway using BEV LiDAR imagery, we create two baselines based on \cite{bai2018} and \cite{homayounfar2018hierarchical}. For the first baseline based on \cite{bai2018} denoted by (DT baseline) , we use the same backbone of the global feature network with additional upsampling and residual layers to only predict the the inverse thresholded DT image at the original image dimension. We threshold the distance transform at 32 pixels on each side of the lane boundary.  Next, we binarize the DT image and skeletonize \cite{suzuki1985topological} to obtain a dense representation of the lane boundaries. This is a very strong baseline since the whole capacity of the backbone network is devoted to predicting the DT image. However, it differs in our method in that we output structured representations of the lane boundaries in the form of polylines in an end-to-end manner that is amenable to an annotator in the loop for correction tasks. We create a second baseline based on \cite{homayounfar2018hierarchical} denoted by (HRAN). The recurrent lane counting module of HRAN architecture has the baked in notion of attending to new lane boundaries from  left to right of the road in the bottom of the image which breaks down for the general case of having new lane boundaries spawning at forks and merges. As such we made this baseline stronger by removing the lane counting module during training and inference and instead providing the ground truth starting points for initialization. Note that our method infers the initial points automatically from the DT image.

\paragraph{Precision/Recall:} We report the precision and recall metric of \cite{wang2016torontocity}. 
Precision is defined as the number of densified predicted polyline points that fall within a threshold of the GT polylines divided by the total length of the predicted polylines. The sums and the divisions are taken over all the images rather than per lane boundary or per image similar to object detection task metrics. Recall is the same but the role of prediction and GT polylines are reversed. We evaluate at 2,3,5, and 10 pixels corresponding to 10, 15, 25 and 50 cm. 
As we can see in Table \ref{tab:baseline}, we outperform both baselines. For DT baseline, we obtain higher precision and recall by a wide margin at 2 and 3 pixels corresponding to 10 and 15. However, this baseline performs better at precision and comparable at recall at 5 and 10 pixels corresponding to 25 and 50 cm. For self driving applications where cm accuracy in HD maps is desirable for the safe navigation of the self driving vehicle, our method proves valuable. For the \cite{homayounfar2018hierarchical} HRAN baseline, where we made it stronger by using GT initial points, we observe that its performance is significantly worse. This is due to HRAN not having an explicit notion of state in the model which prevents the lane boundaries to split at forks and terminate at merges. 

\begin{figure}[t]
	\vspace{-0.5cm}
	\centering
	\setlength{\tabcolsep}{1pt}
	\begin{tabular}{cc}

		\raisebox{15px}{\rotatebox{90}{GT \hspace*{12px} Pr}}		
		\includegraphics[width=0.99\linewidth, trim={0 0 128px 0},clip]{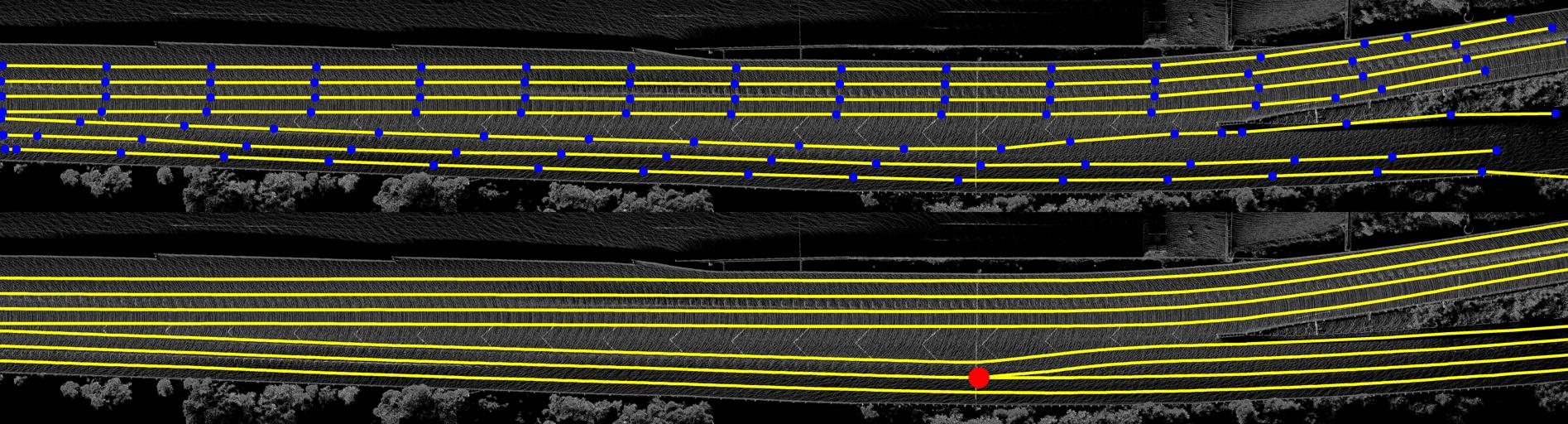}  \\ 
		\raisebox{15px}{\rotatebox{90}{GT \hspace*{12px} Pr}}
		\includegraphics[width=0.99\linewidth, trim={0 0 128px 0},clip]{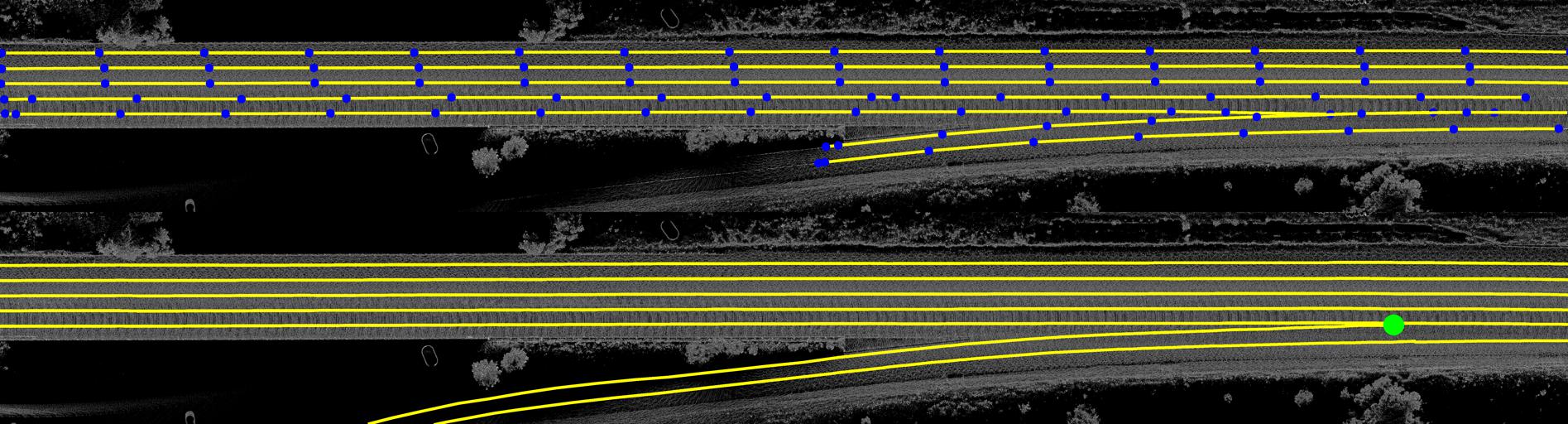}  \\

	\end{tabular}
	\caption{Generalization to a new highway: We train on all the images of one highway and test on another highway in a completely different geographical region to test the generalization of our model. Please zoom.}
	\label{fig:sf}
\end{figure}

\paragraph{Topology:}

For an annotator in the loop to make the least number of clicks, it is desirable that only one predicted polyline segment is assigned to a GT polyline. To assess this, we assign each predicted polyline to the GT lane boundary with the largest intersection at a distance of 20 pixels within the GT. We find that 89\% of GT lane boundaries have the correct topology.

\paragraph{Ablation Study:}

In our ablation studies, we evaluate the importance of the global feature network (G), the Convolutional RNN memory (M) for the headers as well as re-initializing from the DT prediction (R) when we miss drawing lane boundaries. Finally, we evaluate the importance of the notion of state (S) by removing it completely from our model.  That is the node $v_i$ will not have a state component $s_i$ and the position $x_i$ and direction $\theta_i$ will not be conditioned on $s_i$.

To assess the importance of the global feature network, we apply the direction, state and position header directly on the LiDAR image at a local ROI. This way the receptive field of each header is restricted to the attended ROI and lacks a global context of lane boundaries. By comparing rows 1 and 2 of Table \ref{tab:ablation}, we see that just adding the global feature network increases the recall by a wide margin. The precision remains competitive as the model can still do a good job predicting the position of the lane boundaries in a local region. 

Furthermore, we observe in rows 3 and 5 of Table \ref{tab:ablation}, that adding memory and DT recovery to the global feature network, again boosts the performance by a large margin especially in recall. However, we note that the largest gain is obtained from the memory components. If we remove the notion of state and keep all the other components, both precision and recall suffer dramatically as shown in row 4. This highlights the importance of having state in our model. Finally, we note that our full model with all the components has slightly lower precision than the model with no recovery (row 5) and better recall but overall performs the best in terms of F1. We note that these ablations are the average on the test sets of the three cross validation splits.

\paragraph{Qualitative Results:}

Please refer to Figure \ref{fig:qual} for qualitative results. In row 1 we showcase how our model correctly infers the change of topology by spawning two new lane boundaries at a fork. In rows 2 and 3, we demonstrate the behavior of our model at a merge where two lane boundaries become one and either one or both terminate. In row 4. we showcase a failure mode. At the fork, although we have detected a change of topology, the direction header has failed to infer the correct ROI. At the merge, the upper merging lane boundary overlaps a straight one without stopping. Here an annotator has to manually fix these problematic cases rather than drawing everything from scratch.

\paragraph{Generalization to a New Highway:}

The ultimate goal of our method is to enable large scale high definition mapping of highways to in turn facilitate safe self driving at scale. To evaluate the generalization of our method, we train a new model on all the images in the train,val, and test splits of our dataset and we evaluate on a different highway located in a different geographical region. In Table \ref{tab:sf}, we evaluate our model in terms of precision and recall. Although the performance is dropped as expected in comparison to Table \ref{tab:baseline} where we evaluated on the same highway, we still obtain very promising results. In particular, at 2 pixels or 5 cm, we have $57.4\%$ precision and $53.0\%$ recall while at a wider threshold of 10 pixels or 50 cm, we obtain precision and recall values of $88.6\%$ and $80.9\%$ respectively. In Figure \ref{fig:sf}, we demonstrate our predictions on this new highway.

\paragraph{Failure Cases:}

The failure modes consists when a lane boundary is partially not drawn or missed completely and is reflected in our recall metric. In these cases, the annotators have to correct only the problematic cases rather than draw everything from scratch.

%% file: conclusion.tex
\section{Conclusion}

In this paper we have shown how  to draw the lane boundaries of complex highways with many lanes that contain topology changes due to forks and merges.  
Towards this goal, we have formulated the problem as inference in a DAG, where the nodes of the graph encode geometric and topological properties of the local regions of the lane boundaries. We  also derived a simple yet effective greedy algorithm that allow us to infer  the DAG topology (i.e., nodes and edges) for each region. 
We have demonstrated the effectiveness of our approach on  two major North American Highways in two different states and show high accuracy in terms of the line drawing and the topology recovered.  
We plan to extend our approach to utilize also  cameras that are typically available onboard  self driving vehicles to further improve the performance.

%% file: supp_model.tex

In Sec. \ref{sec:net} we present the components of our model in detail. In Sec. \ref{sec:train}, we discuss the training details. Finally, we showcase more qualitative examples in Sec. \ref{sec:qual}.

\section{Network Architecture}
\label{sec:net}
\begin{figure*}[h]

	\centering

		\includegraphics[width=0.5\linewidth]{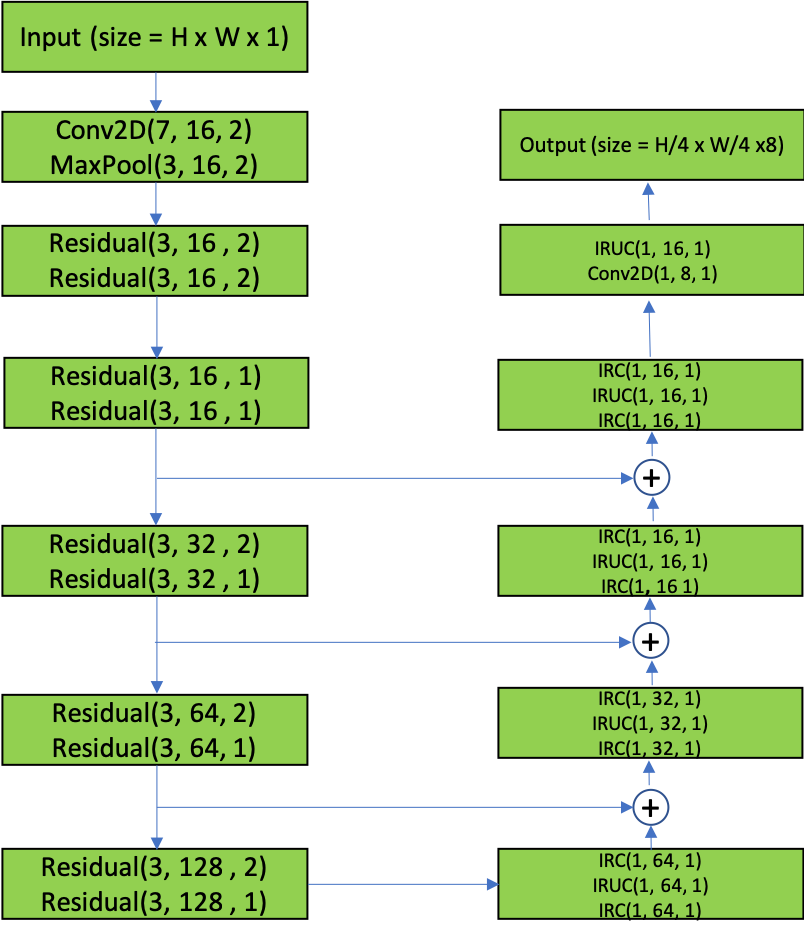}  \\

	\caption{Global Feature Network}
	\label{fig:glob_model}
\end{figure*}

\paragraph{Global Feature Network:} In Fig. \ref*{fig:glob_model}, we visualize a diagram of the global feature network. In the following:
\begin{itemize}

\item {\bf Conv2D(kernel size, filters, stride)} is a 2D convolution layer.
\item Each residual {\bf Residual(kernel size, Number of filters, stride)} layer is comprised of two blocks of instance normalization, ReLU nonlineary and Conv2d(kernel size, filters, stride (or 1)). Note that some residual layers perform downsampling by setting the stride of the first convolution layer to two while keeping the stride of the second convolution as one.    
\item {\bf IRC(kernel size, filters, stride)} corresponds to instance normalization, ReLU followed by a Conv(kernel size, filters, stride) while {\bf IRUC(kernel size, filters, stride)} has an extra nearest neighbor upsampling layer after the ReLU.

\end{itemize}

\begin{figure*}[h]
	
	\centering
	
	\includegraphics[width=0.99\linewidth]{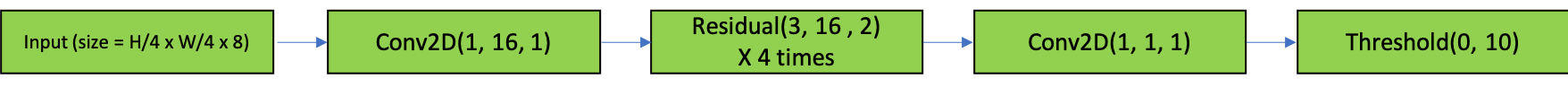}  \\

	\caption{Distance Transform Header}
	\label{fig:dt_header}
\end{figure*}

\paragraph{Distance Transform Header:} We input the features obtrained from the Global Feature Network of Fig. \ref{fig:glob_model} to the distance transform (DT) header of Fig. \ref{fig:dt_header}. The GT places the value of 8 on the lane boundaries and decreases as we move away. As such we clip the output of the DT header to be between 0 and 10.

\paragraph{Direction Header:} We crop a rotated RoI of size $h \times w$ along the lane boundary from the output of the global feature network and concatenate with the one-hot encoding of the previous inferred state to obtain a tensor with 11 channels. In the following:

\begin{itemize}
	\item {\bf ConvRNN(kernel size, filters, stride)} is a vanilla RNN that replaces matrix multiplication with convolutions and the $\tanh$ non-linearity with ReLU. 
	\item {\bf Linear(input size, output size)} is a simple linear linear
\end{itemize}

\begin{figure*}[h]
	
	\centering
	
	\includegraphics[width=0.50\linewidth]{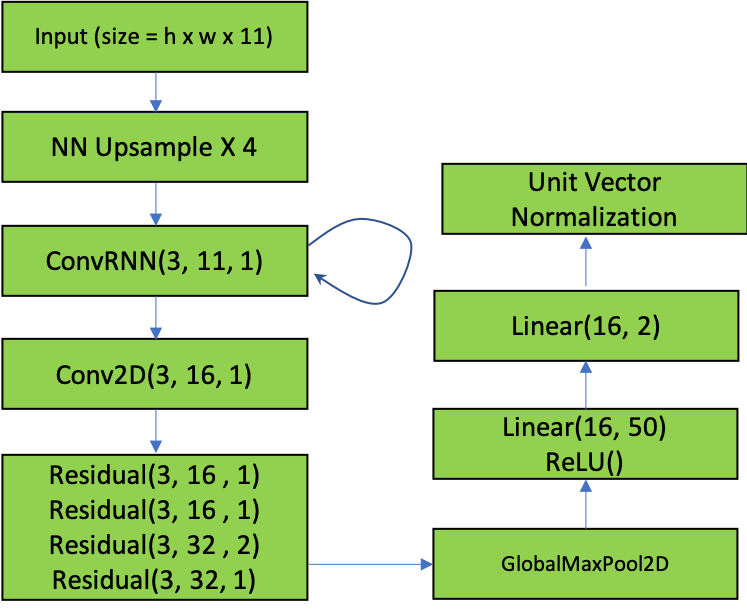}  \\

	\caption{Direction Header}	
	\label{fig:dir_header}	
\end{figure*}

\paragraph{State Header:} The state header has the exact same architecture as of the direction header. However, we replace the global max pooling with global average pooling.

\begin{figure*}[h]
	
	\centering
	
	\includegraphics[width=0.50\linewidth]{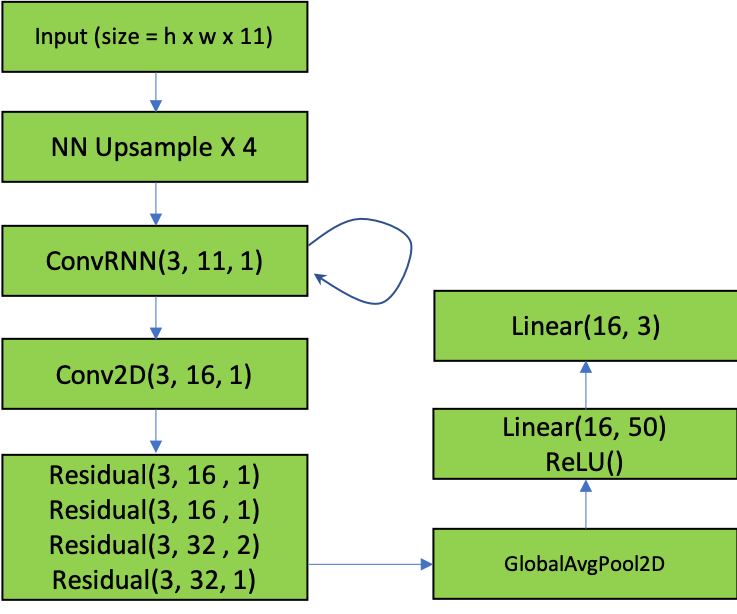}  \\
	
	\caption{State Header}	
	\label{fig:state_header}	
\end{figure*}

\begin{figure*}[h]
	
	\centering
	
	\includegraphics[width=0.50\linewidth]{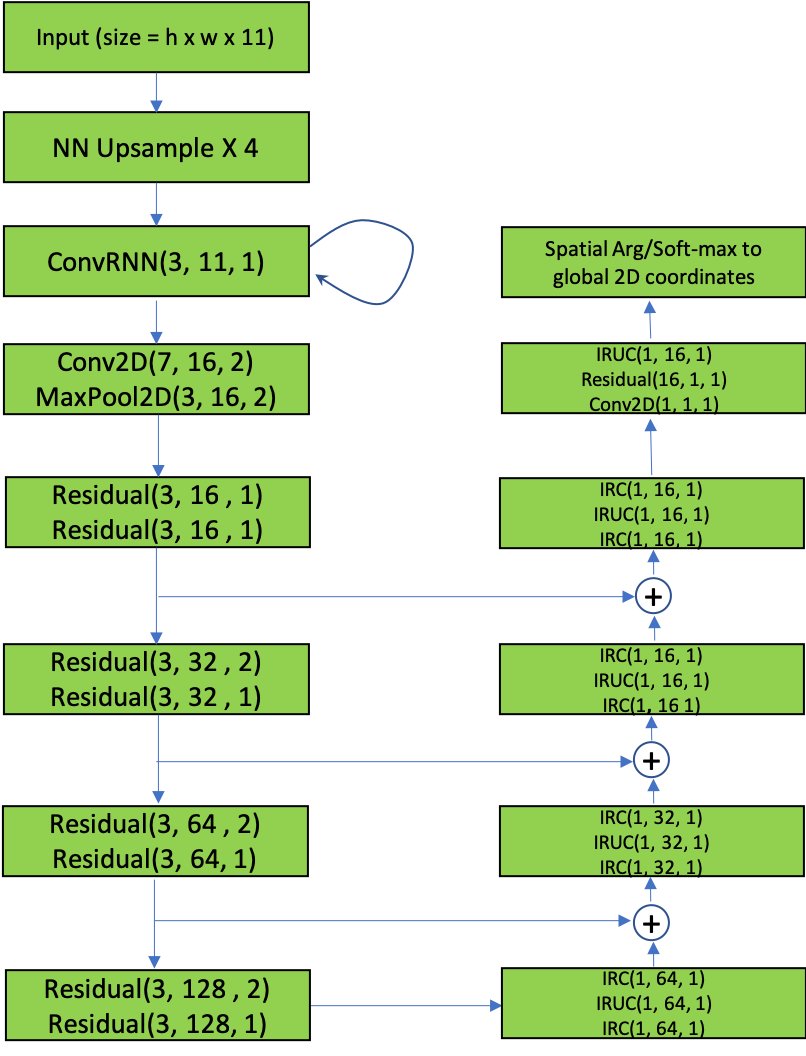}  \\
	
	\caption{Position Header}	
	\label{fig:pos_header}	
\end{figure*}

\paragraph{Position Header:} For the position header, we employ a convolutional RNN coupled with an encoder decoder network as outlined in Fig. \ref{fig:pos_header}.

%% file: supp_training.tex

\section{Training Details}
\label{sec:train}
In this section, we describe the training implementation details. For learning, we optimize the following multi-task objective:

\begin{align}
 \mathcal{L}(\theta)  = \lambda_1 \ell_{chamfer}(\theta) + \lambda_2 \ell_{cosine}(\theta) + \lambda_3 \ell_{focal}(\theta) + \lambda_4\ell_{dt}(\theta)
\end{align} 
where $\theta$ are the parameters of our neural network. We utilize the validation set to obtain a best set of hyperparameters as $\lambda_1 = 1, \lambda_2=100$ and $\lambda_3=\lambda_4=10$. Since each image is of dimension $1200\times 8000$ pixels and our model comprises of three convolutional RNNs, we are able to fit only one example in an NVIDIA 1080 Ti GPU during training.  We trained our model on 8 GPUs using the distributed training environment Horovod  for 12 hours. We used the AMSgrad  optimizer with the learning rate of 0.001. We choose the final model based on the lowest precision and recall on the validation set.

During training, we provide all the GT initial vertices $\{v_{\text{init}} = (\theta_{\text{init}}, x_{\text{init}}, s_{\text{init}})\}$ with a small Gaussian perturbation on the initial positions  $x_{\text{init}}$. Next, given a parent node $v_{\P(i)}$,  we predict the components of the next vertex $\{v_{i} = (\theta_{i}, x_{i}, s_{i})\}$. We replace all the $\arg\max$ operators with $softmax$. The ground truth for  $\theta_{i}$ is generated on the fly by projecting $x_{\P(i)}$, the anchor for the current Region of interest (RoI), onto the closest lane boundary and computing the turning function at that point. The GT for the categorical random variable $s_i$ is obtained also on the fly by determining whether the current RoI contains a fork or a terminating point and setting the GT accordingly. If neither topological change falls within the RoI, we set the GT to be the normal state.

While training, we use the predicted direction $\theta_{i}$ anchored at $x_{\P(i)}$ to obtain the next RoI unless $x_{\P(i)}$ is 30 pixels away from the lane boundary, in which case we use the GT direction. Moreover, we concatenate the GT states to each header during training. For training the convolutional RNN components of the direction, state and position header, we backpropagate 5 timesteps through time to avoid memory issues. We run the RNN until either the vertex position falls outside the image, the rotated RoI contains a GT terminate state or the RNN runs for 30 timesteps.

At test time, we use the predicted initial vertices, the directions and the states to infer the state of the DAG. 

%% file: qual.tex

\section{Qualitative Results}
\label{sec:qual}
\begin{figure*}[h]
	\centering
	\setlength{\tabcolsep}{1pt}
	\begin{tabular}{cc}

		\raisebox{10px}{\rotatebox{90}{GT \hspace*{2em} Pr \hspace*{1.5em} Inp}}
		
		\includegraphics[width=0.99\linewidth]{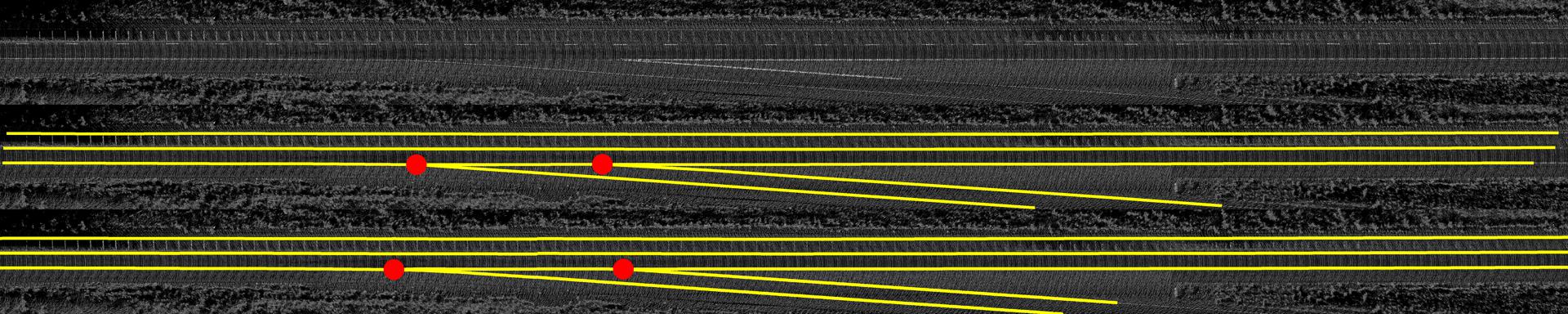}  \\

		\raisebox{10px}{\rotatebox{90}{GT \hspace*{2em} Pr \hspace*{1.5em} Inp}}
		
		\includegraphics[width=0.99\linewidth]{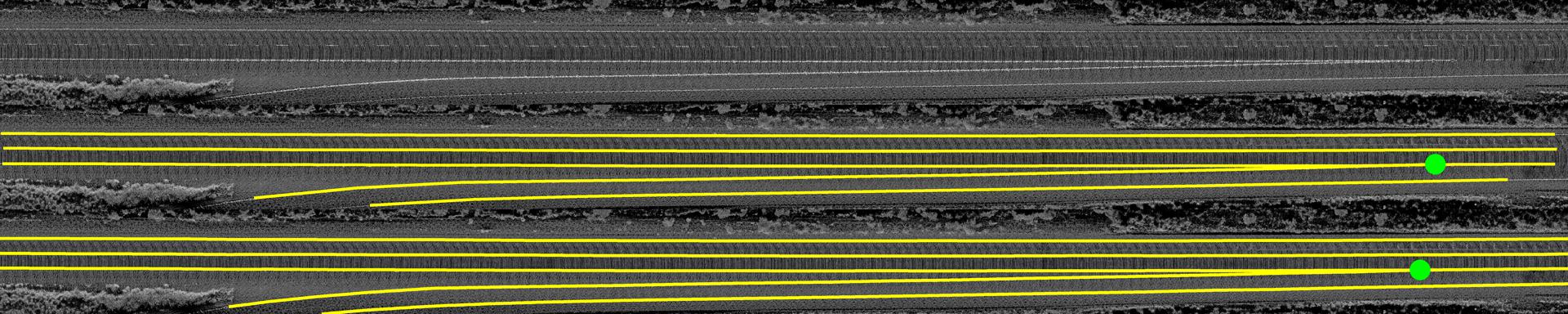}  \\

	\end{tabular}
	\caption{Qualitative results where we showcase the Input Lidar Image (Inp), the predictions (Pr) and the Ground Truth (GT).}
	\label{fig:qual1}
\end{figure*}

\begin{figure*}[h]
	\centering
	\setlength{\tabcolsep}{1pt}
	\begin{tabular}{cc}

		\raisebox{2.5em}{\rotatebox{90}{GT \hspace*{5em} Pr \hspace*{5em} Inp}}
		\includegraphics[width=0.99\linewidth]{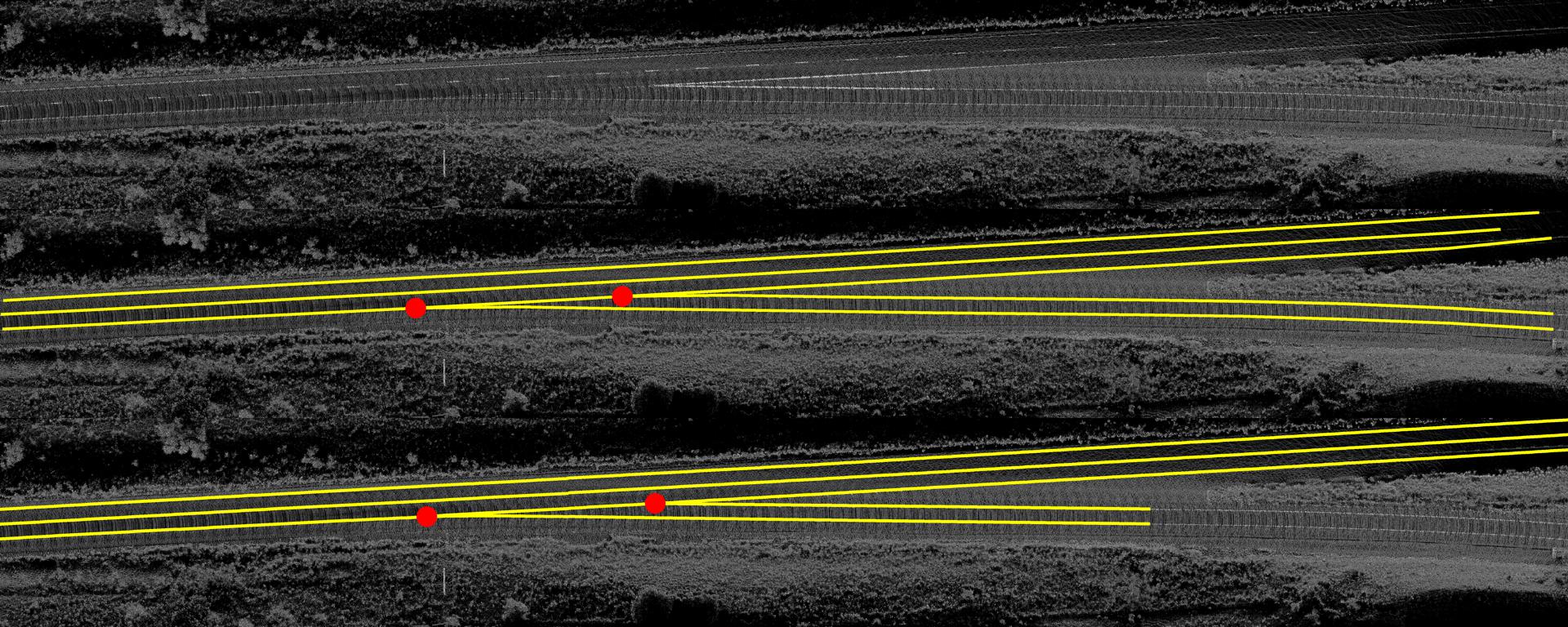}  \\

		\raisebox{2.5em}{\rotatebox{90}{GT \hspace*{5em} Pr \hspace*{5em} Inp}}
		\includegraphics[width=0.99\linewidth]{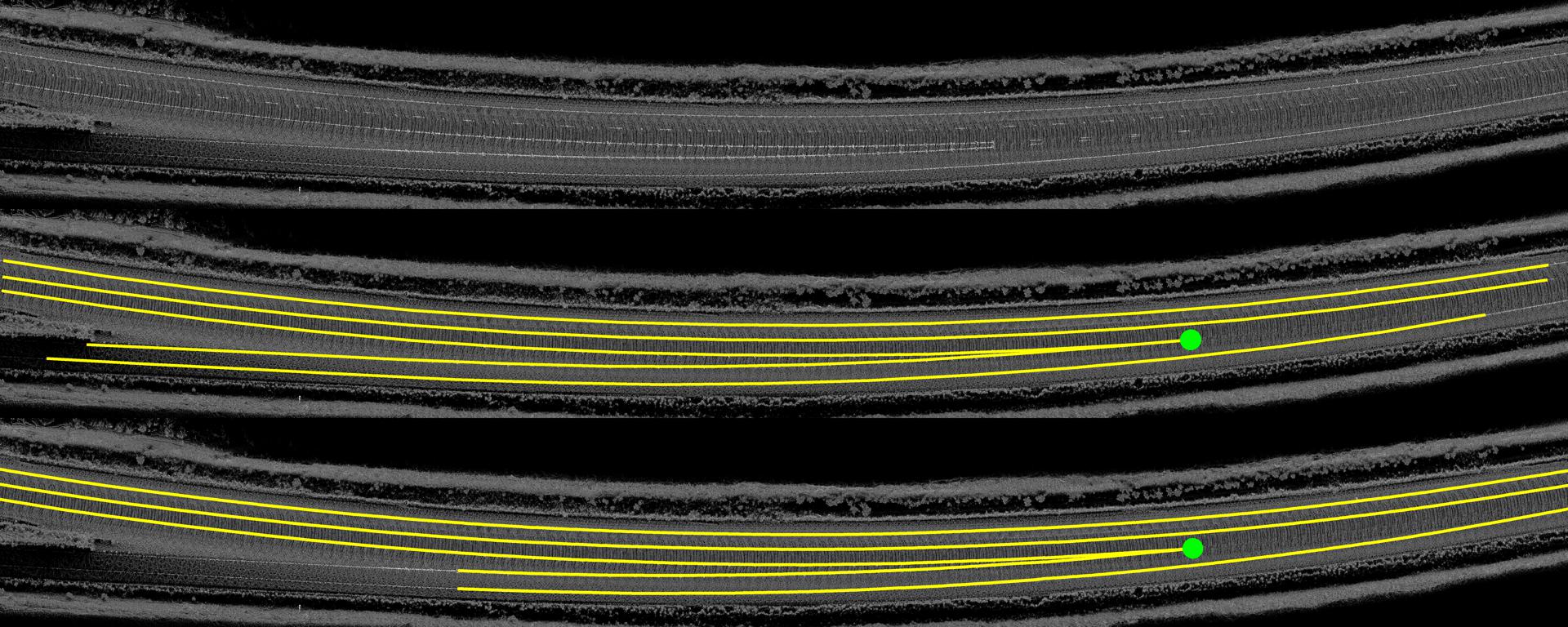}  \\
		
		\raisebox{2.5em}{\rotatebox{90}{GT \hspace*{5em} Pr \hspace*{5em} Inp}}
		\includegraphics[width=0.99\linewidth]{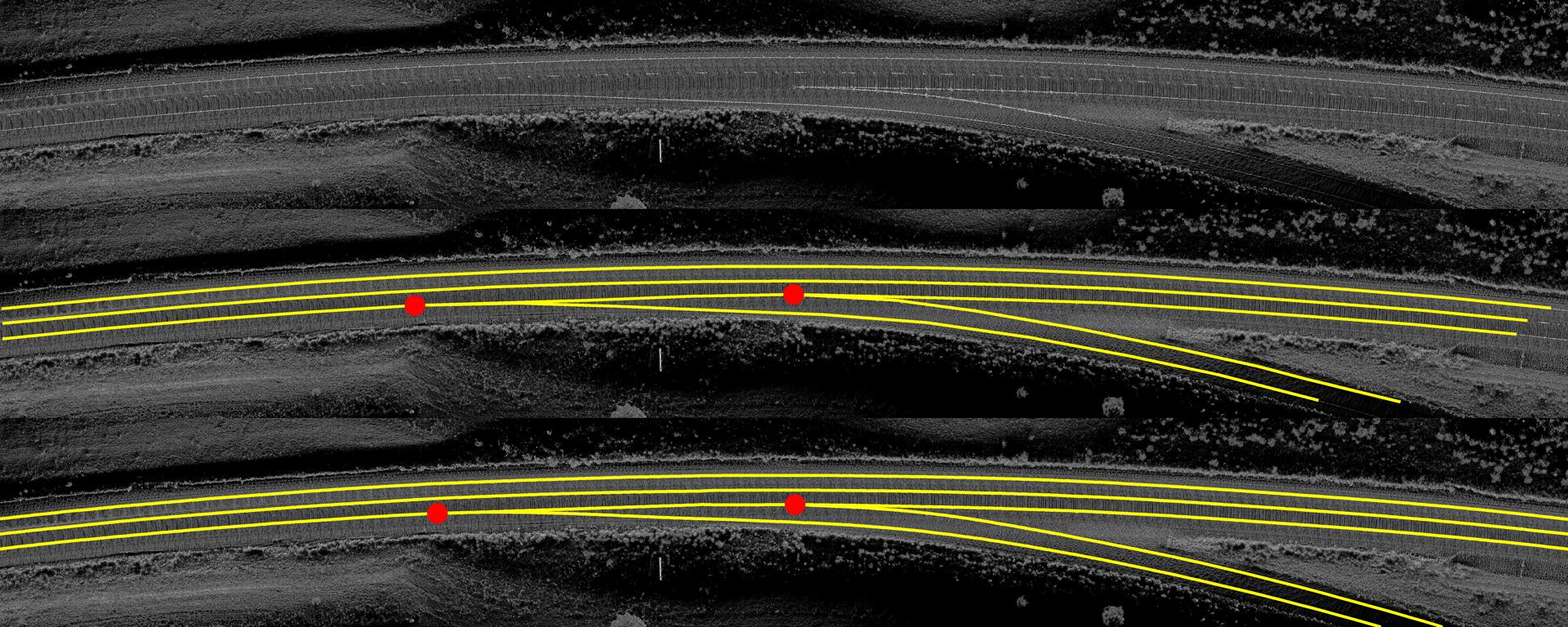}  \\

	\end{tabular}
	\caption{Qualitative results where we showcase the Input Lidar Image (Inp), the predictions (Pr) and the Ground Truth (GT).}
	\label{fig:qual2}
\end{figure*}

\begin{figure*}[h]
	\centering
	\setlength{\tabcolsep}{1pt}
	\begin{tabular}{cc}

		\raisebox{2.5em}{\rotatebox{90}{GT \hspace*{5em} Pr \hspace*{5em} Inp}}
		\includegraphics[width=0.99\linewidth]{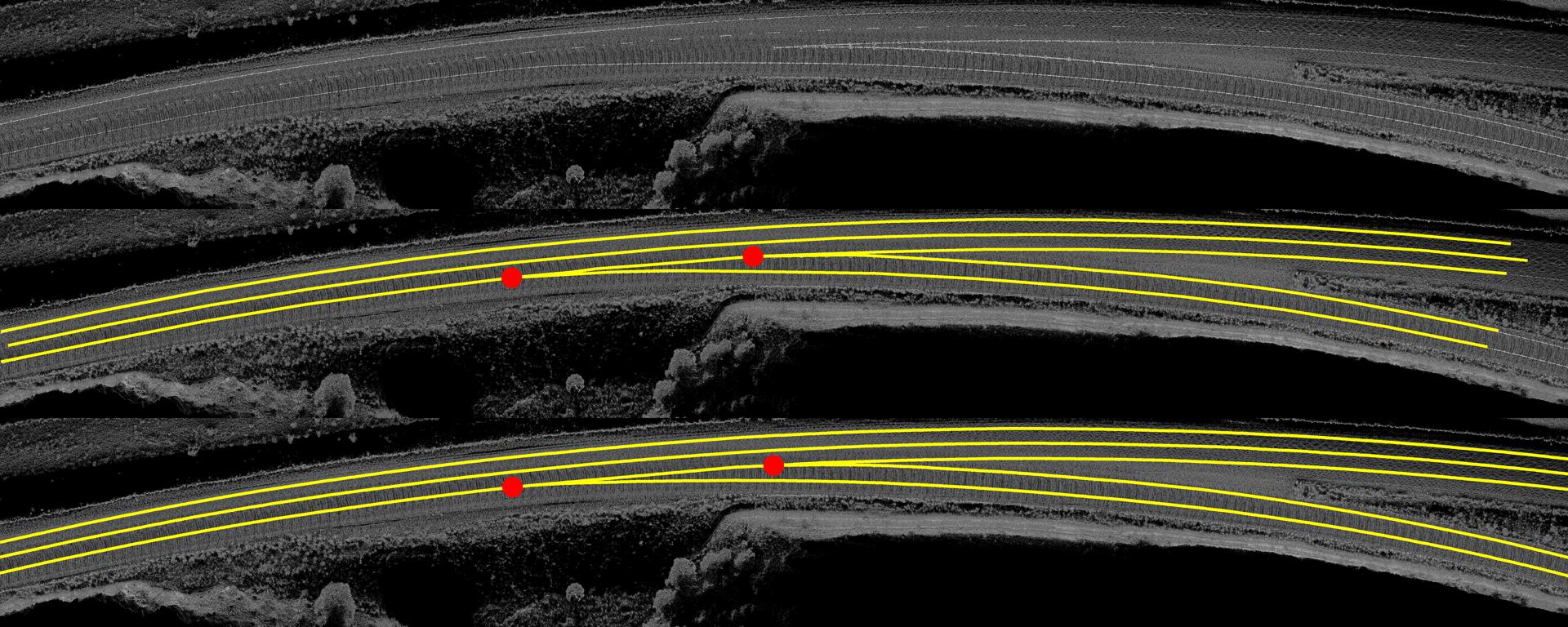}  \\
		
		\raisebox{2.5em}{\rotatebox{90}{GT \hspace*{5em} Pr \hspace*{5em} Inp}}
		\includegraphics[width=0.99\linewidth]{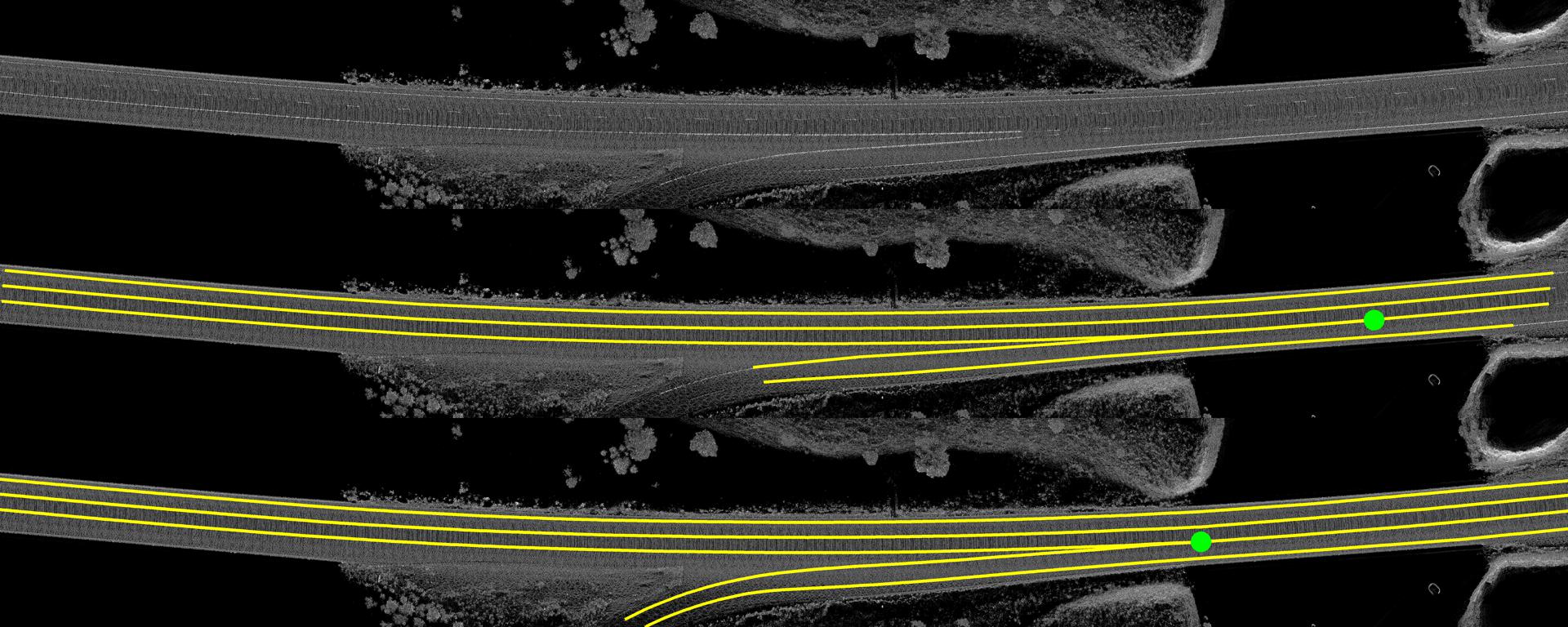}  \\

		\raisebox{2.5em}{\rotatebox{90}{GT \hspace*{5em} Pr \hspace*{5em} Inp}}
		\includegraphics[width=0.99\linewidth]{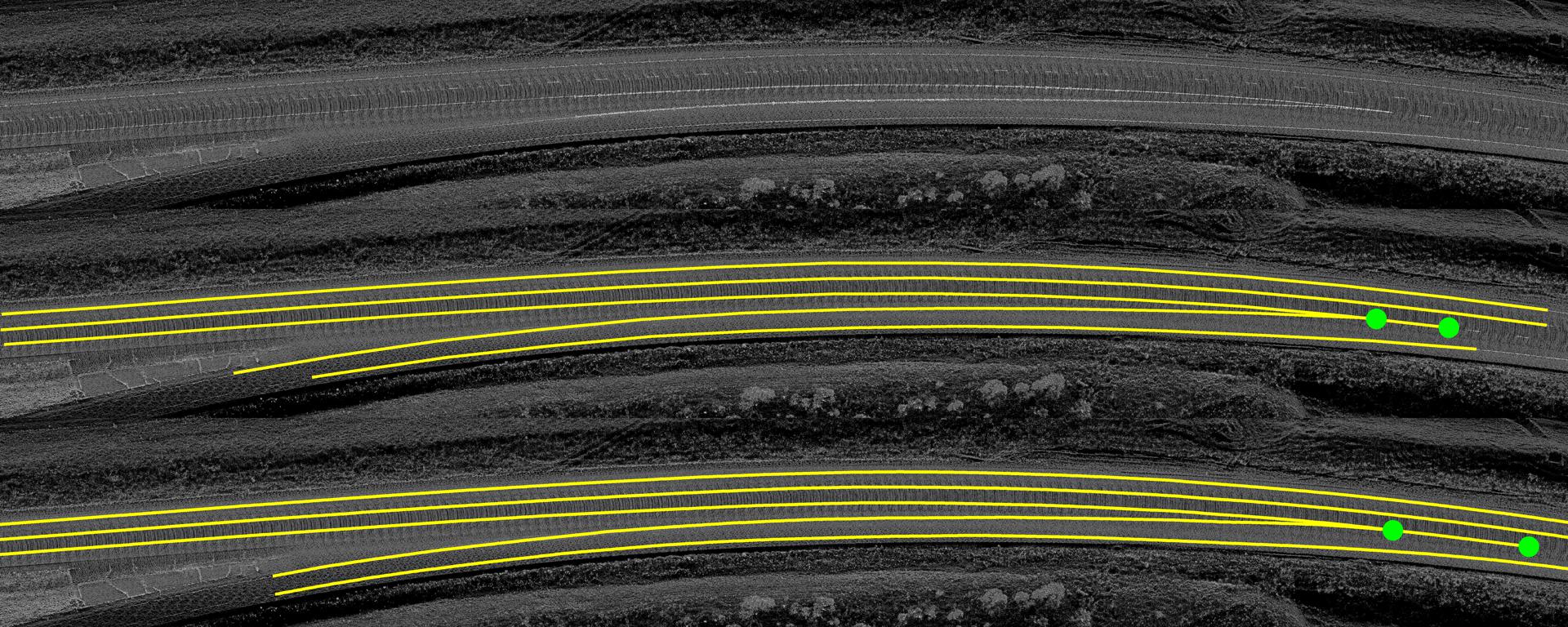}  \\

	\end{tabular}
	\caption{Qualitative results where we showcase the Input Lidar Image (Inp), the predictions (Pr) and the Ground Truth (GT).}
	\label{fig:qual3}
\end{figure*}

\begin{figure*}[h]
	\centering
	\setlength{\tabcolsep}{1pt}
	\begin{tabular}{cc}

		\raisebox{2.5em}{\rotatebox{90}{GT \hspace*{5em} Pr \hspace*{5em} Inp}}
		\includegraphics[width=0.99\linewidth]{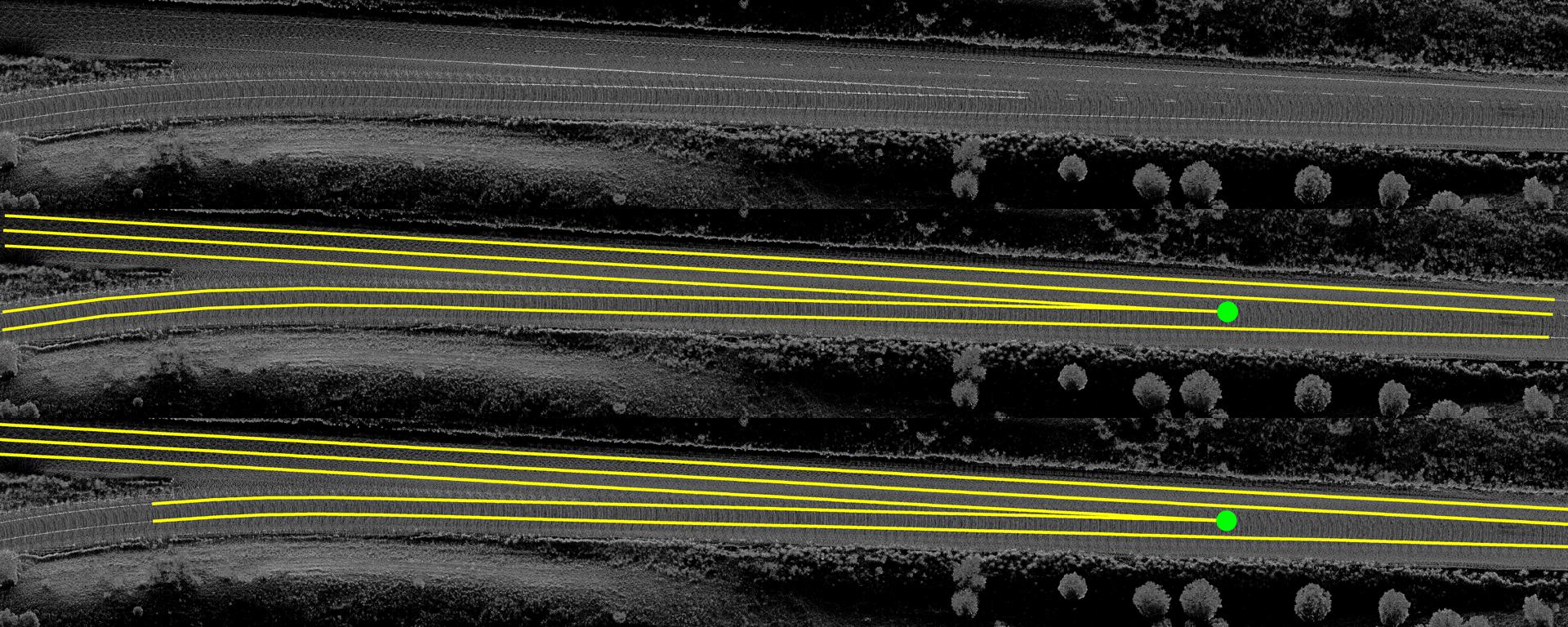}  \\
		
		\raisebox{0.7em}{\rotatebox{90}{GT \hspace*{1.8em} Pr \hspace*{1.7em} Inp}}
		\includegraphics[width=0.99\linewidth]{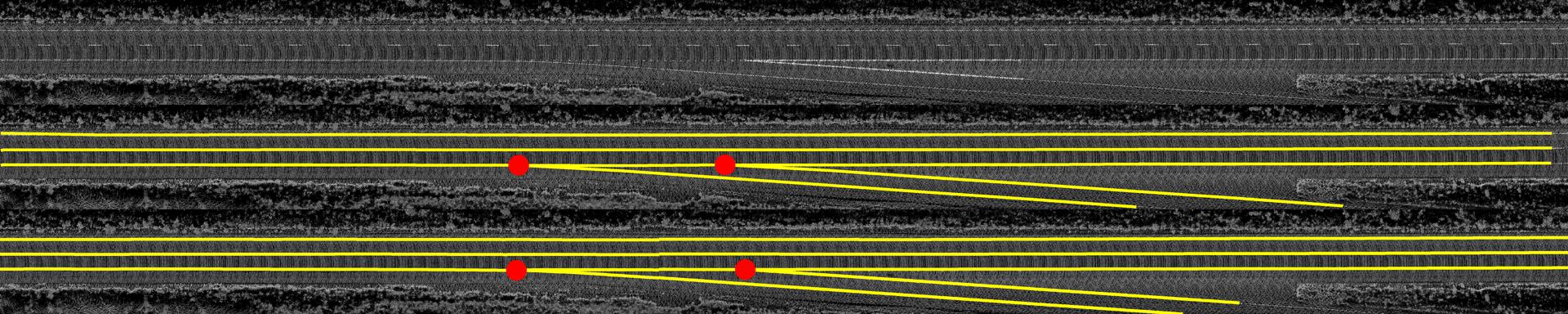}  \\
		
		\raisebox{0.8em}{\rotatebox{90}{GT \hspace*{1.8em} Pr \hspace*{1.7em} Inp}}
		\includegraphics[width=0.99\linewidth]{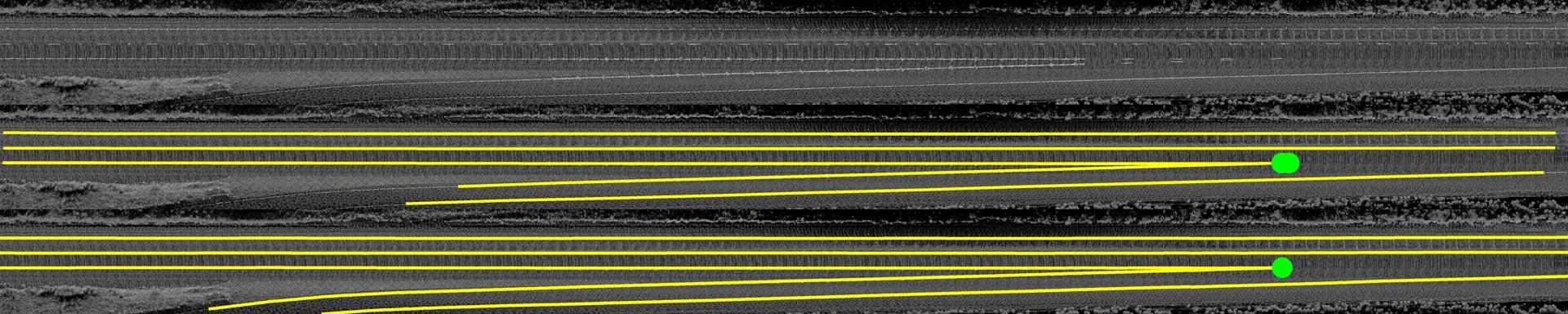}  \\
		
		\raisebox{0.8em}{\rotatebox{90}{GT \hspace*{1.8em} Pr \hspace*{1.7em} Inp}}
		\includegraphics[width=0.99\linewidth]{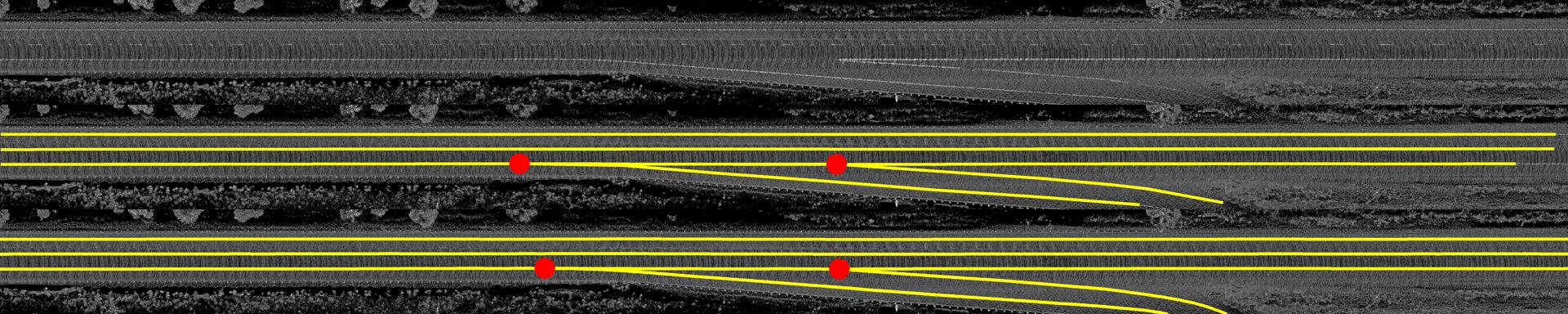}  \\

		\raisebox{0.8em}{\rotatebox{90}{GT \hspace*{1.8em} Pr \hspace*{1.7em} Inp}}
		\includegraphics[width=0.99\linewidth]{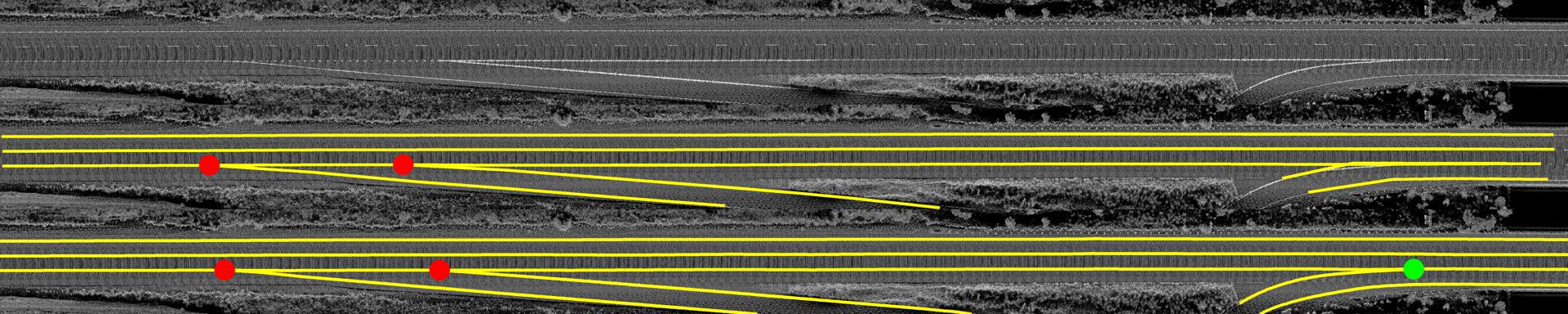}  \\

	\end{tabular}
	\caption{Qualitative results where we showcase the Input Lidar Image (Inp), the predictions (Pr) and the Ground Truth (GT).}
	\label{fig:qual4}
\end{figure*}

\begin{figure*}[t]
	\centering
	\setlength{\tabcolsep}{1pt}
	\begin{tabular}{cc}

		\raisebox{2.5em}{\rotatebox{90}{GT \hspace*{5em} Pr \hspace*{5em} Inp}}
		\includegraphics[width=0.99\linewidth]{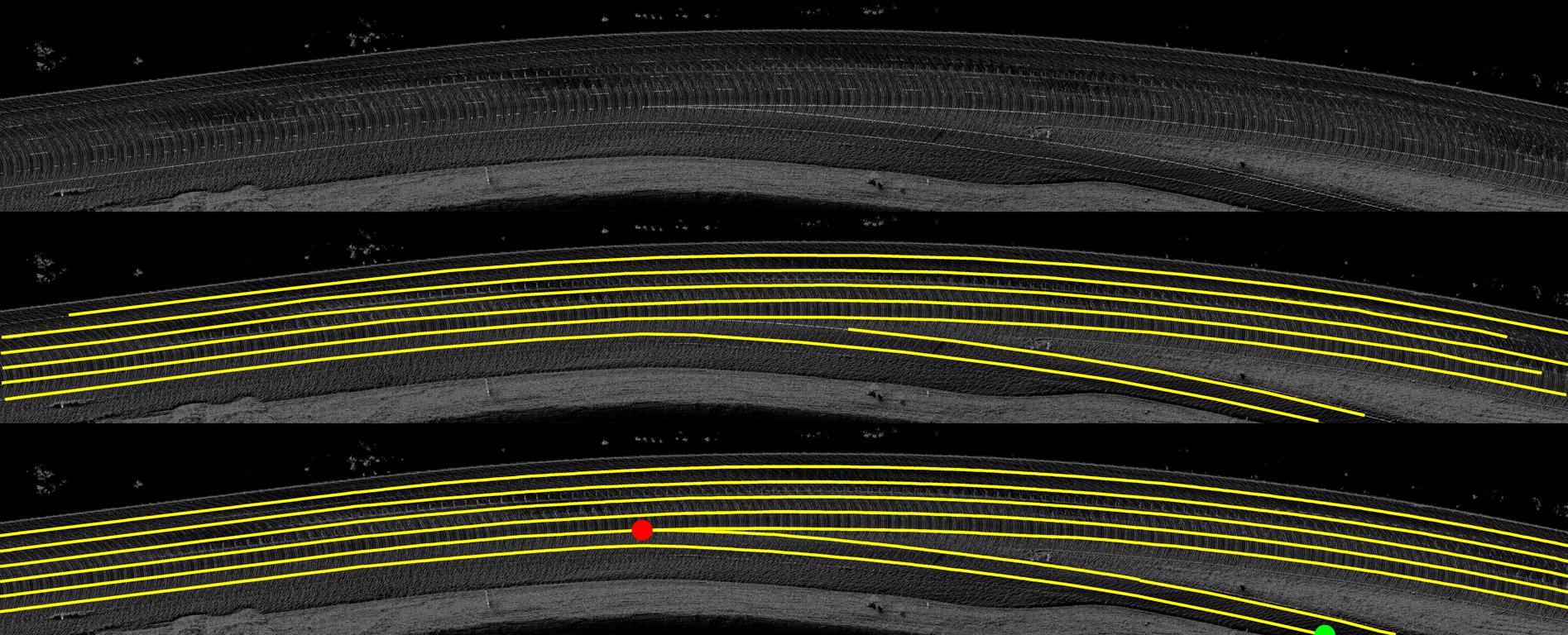}  \\
		
		\raisebox{2.5em}{\rotatebox{90}{GT \hspace*{5em} Pr \hspace*{5em} Inp}}
		\includegraphics[width=0.99\linewidth]{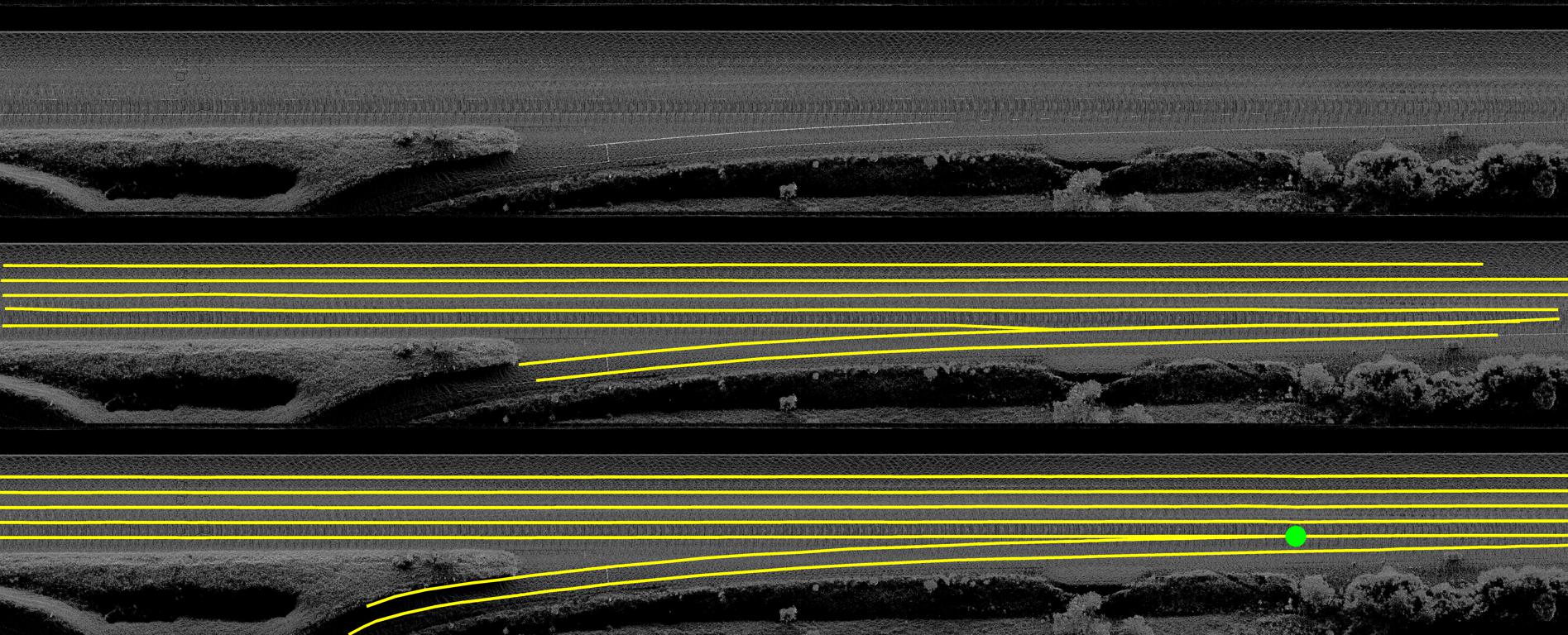} \\
		
		\raisebox{2.5em}{\rotatebox{90}{GT \hspace*{5em} Pr \hspace*{5em} Inp}}
		\includegraphics[width=0.99\linewidth]{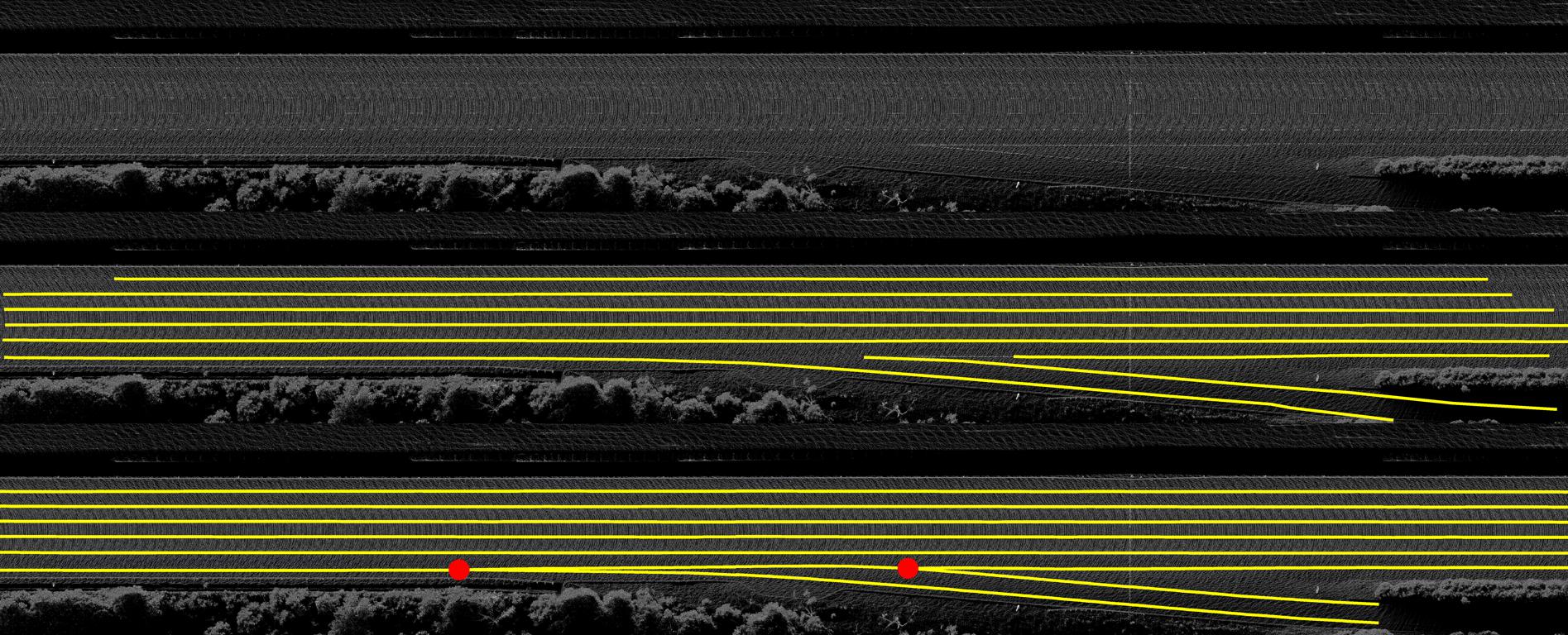}

	\end{tabular}
	\caption{Generalization from training on one highway and testing on another 1000 km away. we showcase the Input Lidar Image (Inp), the predictions (Pr) and the Ground Truth (GT).}
	\label{fig:gen}
\end{figure*}

\begin{figure*}[t]
	\centering
	\setlength{\tabcolsep}{1pt}
	\begin{tabular}{cc}

		\raisebox{2.5em}{\rotatebox{90}{GT \hspace*{5em} Pr \hspace*{5em} Inp}}
		\includegraphics[width=0.99\linewidth]{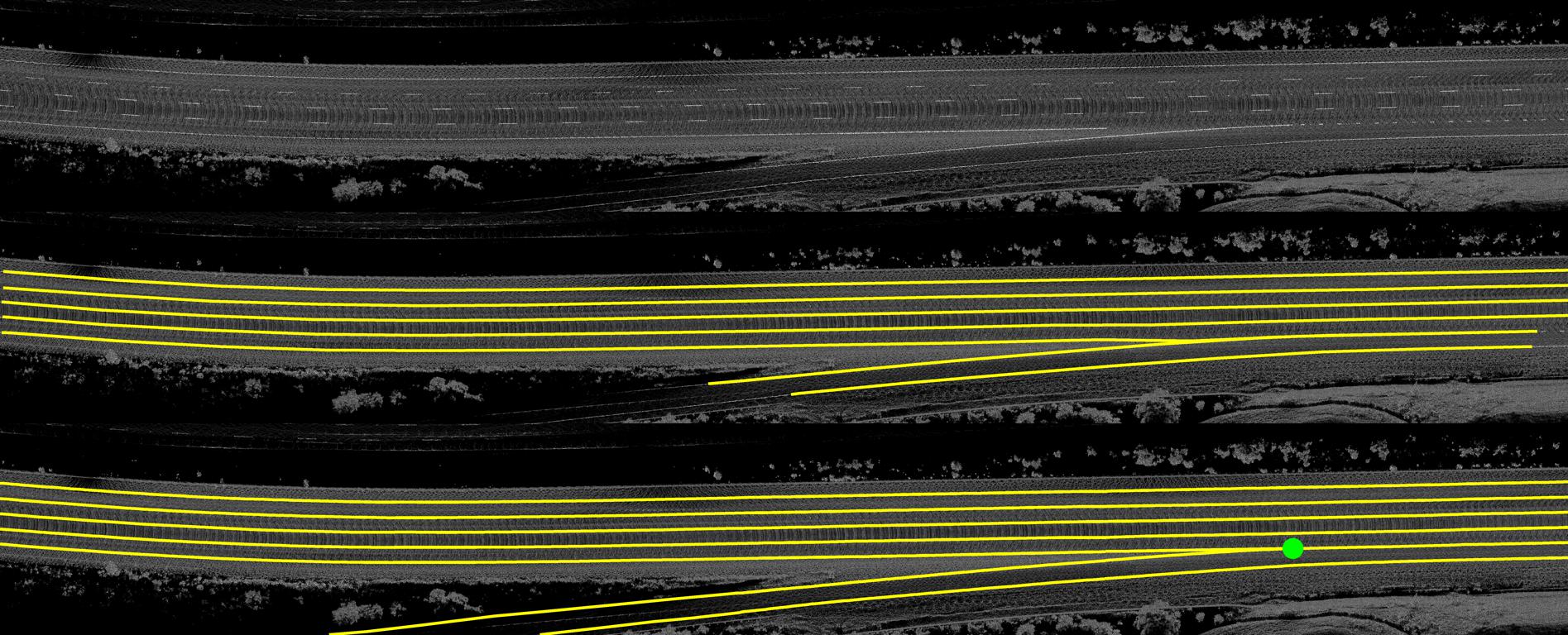}  \\
		
		\raisebox{2.5em}{\rotatebox{90}{GT \hspace*{5em} Pr \hspace*{5em} Inp}}
		\includegraphics[width=0.99\linewidth]{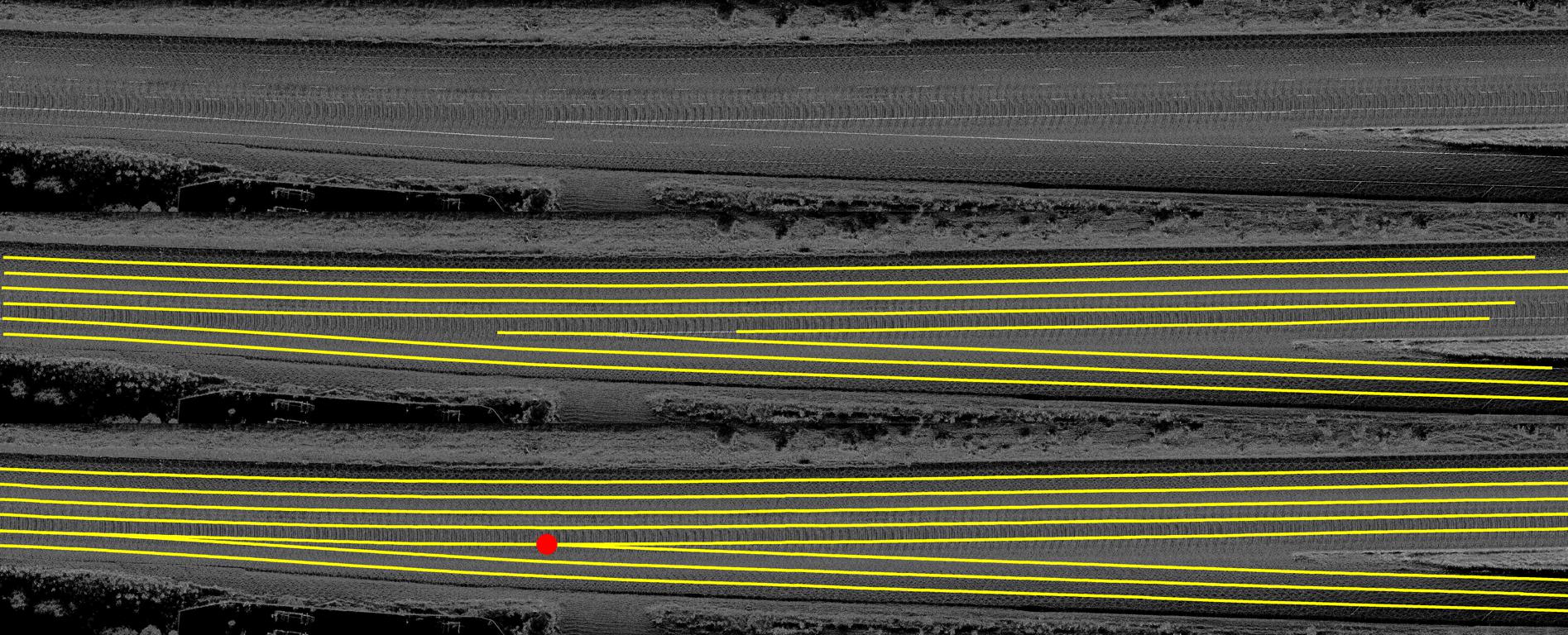} \\
		
		\raisebox{2.5em}{\rotatebox{90}{GT \hspace*{5em} Pr \hspace*{5em} Inp}}
		\includegraphics[width=0.99\linewidth]{./supp_sf_figs/801_29.jpg}

	\end{tabular}
	\caption{Generalization from training on one highway and testing on another 1000 km away. we showcase the Input Lidar Image (Inp), the predictions (Pr) and the Ground Truth (GT).}
	\label{fig:gen2}
\end{figure*}

In Figs. \ref{fig:qual1}, \ref{fig:qual2}, \ref{fig:qual3} and \ref{fig:qual4} we demonstrate our predictions on the first highway. In Fig. \ref{fig:gen} and \ref{fig:gen2} we showcase the results of training only on one highway and testing on another highway 1000 km away.

